\documentclass{tlp}

\usepackage{amsmath}
\usepackage{graphicx}
\usepackage{array}
\usepackage{multirow}
\usepackage{caption} 
\usepackage{subcaption}
\usepackage{url}
\usepackage{xcolor}

\usepackage{natbib}
\renewcommand\harvardyearright[1]{.}

\setcitestyle{author,open={(}, authoryear ,close={)}} 

\newtheorem{example}{Example}

\begin{document}

\lefttitle{H.M.Qureshi and W.Faber}

\jnlPage{1}{8}
\jnlDoiYr{2021}
\doival{10.1017/xxxxx}

\title[Evaluating Datalog Tools for Meta-reasoning over OWL 2 QL]{Evaluating Datalog Tools for Meta-reasoning over OWL 2 QL}

\begin{authgrp}
\author{\sn{Qureshi} \gn{Haya Majid}}
\affiliation{University of Klagenfurt}
\author{\sn{Faber} \gn{Wolfgang}}
\affiliation{University of Klagenfurt}
\end{authgrp}

\history{\sub{xx xx xxxx;} \rev{xx xx xxxx;} \acc{xx xx xxxx}}

\maketitle

\begin{abstract}
  Metamodeling is a general approach to expressing knowledge about classes and properties in an ontology. It is a desirable modeling feature in multiple applications that simplifies the extension and reuse of ontologies. Nevertheless, allowing metamodeling without restrictions is problematic for several reasons, mainly due to undecidability issues. Practical languages, therefore, forbid classes to occur as instances of other classes or treat such occurrences as semantically different objects. Specifically, meta-querying in SPARQL under the Direct Semantic Entailment Regime (DSER) uses the latter approach, thereby effectively not supporting meta-queries. However, several extensions enabling different metamodeling features have been proposed over the last decade. This paper deals with the Metamodeling Semantics (MS) over OWL 2 QL and the Metamodeling Semantic Entailment Regime (MSER), as proposed in \cite{lenzerini2015higher} and \cite{lenzerini2020metaquerying,cima2017sparql}. A reduction from OWL 2 QL to Datalog for meta-querying was proposed in \cite{cima2017sparql}.  In this paper, we experiment with various logic programming tools that support Datalog querying to determine their suitability as back-ends to MSER query answering. These tools stem from different logic programming paradigms (Prolog, pure Datalog, Answer Set Programming, Hybrid Knowledge Bases). Our work shows that the Datalog approach to MSER querying is practical also for sizeable ontologies with limited resources (time and memory). This paper significantly extends \cite{qureshi2021evaluation} by a more detailed experimental analysis and more background. Under consideration in Theory and Practice of Logic Programming (TPLP).
\end{abstract}


\section{Introduction}
The OWL ontology language (see for instance \cite{horrocks2003shiq,RePEc:spr:ihichp:978-3-540-92673-3_4}) provides a powerful data modeling language and automated reasoning abilities based on Description Logics (DL) (see for example \cite{baader2003description}). In the current version, OWL 2, several profiles exist that restrict the language features for limiting complexity of reasoning and easing implementations. In this article, we focus on the OWL 2 QL ontology language (see \cite{motik2009owl}), which is geared towards Ontology-Based Data Access (OBDA) \cite{eiter2012reasoning}, that is, querying a huge amount of pre-existing data residing in possibly external information systems. OWL 2 QL is based on the Description Logic {\em DL-Lite$_{R}$} (see \cite{calvanese2007tractable}) and the tractability of main reasoning tasks with respect to the size of the data is guaranteed, see \cite{motik2009owl}. This is important, as the involved data can be very extensive in real-world applications. 

An issue with ontologies that has gained momentum in recent years is meta-modeling and metaquerying, see for example \cite{motik2007properties,brasileiro2016expressive}. Metamodeling refers to higher-level abstractions in ontologies, where classes, called metaclasses, can have other intensional predicates, i.e., classes, properties or datatypes, as instances, and properties can have as instances pairs including intensional predicates as well. Metaqueries are queries that range over intensional predicates and not only over individuals.

The language OWL 2, and therefore also OWL 2 QL, syntactically allows for meta-modeling employing punning\footnote{\url{http://www.w3.org/2007/OWL/wiki/Punning}}. With punning, the same name can be used for ontology elements of different types (most notably, classes and individuals). However, the standard semantics of OWL 2, the Direct Semantics (DS) \cite{horrocks2012owl}, treats punning in a way that is not intended by meta-modeling. The reason is that DS imposes a sharp distinction between the entity types, interpreting the differently typed occurrences of the same name necessarily as different entities.

The following example, adapted from \cite{guizzardi2015towards}, shows the use of meta-modeling in OWL 2 QL and how it affects query answering under DS.

\begin{example}\label{ex:goldeneagle}
Consider the modeling of biological species, stating that all  GoldenEagles are  Eagles,  all Eagles are  Birds, and that Harry is an instance of GoldenEagle, which by inference is also an instance of Eagle and Birds.
\begin{quote}
\centering \textit{SubClassOf(:Eagle :Birds)}\\
\textit{SubClassOf(:GoldenEagle :Eagle)}\\
\textit{ClassAssertion(:GoldenEagle :Harry)}
\end{quote}
However, in the species domain one can also articulate expressions about species themselves, such as  “GoldenEagle is listed in the IUCN Red List of endangered species”. This states that GoldenEagle as a whole class is an endangered species. This is different from specific properties of species and also does not state anything about individuals, such as Harry. To formally model this statement, we can declare GoldenEagle to be an instance of the new class EndangeredSpecies.

\begin{quote}
\centering \textit{ClassAssertion(:EndangeredSpecies :GoldenEagle)}
\end{quote}

Asserting GoldenEagle to be of type EndangeredSpecies is concise and clear. Note that making GoldenEagle a subclass of EndangeredSpecies would not be correct, as it results in incorrect conclusions like ``Harry is an EndangeredSpecies''. In the assertion above, EndangeredSpecies is a metaconcept for GoldenEagle.  
\end{example}

Despite the syntactic support of punning offered by OWL 2 QL, DS prevents extracting some information from the knowledge base by means of metaquerying. For instance, querying ``Is Harry an individual of a species that is endangered?'' over the ontology of Example~\ref{ex:goldeneagle} is not really possible.

Let us consider this in a bit more detail.
SPARQL (see, for example \cite{perez2009semantics}) is the de-facto standard ontology query language that supports  Basic Graph Patterns (BGP). Simple conjunctive queries are allowed in the language, however, there are type restrictions to be met in DS. Suppose that one formulates a metaquery, for simplicity denoted as a conjunctive query in the style of relational calculus:
\begin{quote}
\centering $\{(z) \mid \exists y\> ES(y) \land y(z) \land Lives\_in(z,CPZ)\}$
\end{quote}
The intention is to retrieve all individuals that live in Central Park Zoo (CPZ) and are instances of a species that is endangered. The corresponding SPARQL query would be:
\[
\begin{array}{l}
  
SELECT\> ?z\> WHERE  \>\{ \\
    \indent  \indent  \indent ?y \> a \> ES .\\
    \indent  \indent  \indent ?z \> a \> ?y .\\
    \indent  \indent  \indent ?z \> Lives\_in \> CPZ \\
 \}\\
\end{array}
\]

Unfortunately, this query is not allowed under the {\em Direct Semantics Entailment Regime (DSER)} \cite{glimm2011using}, which is the logical underpinning for SPARQL queries for interpreting the logical theories corresponding to OWL 2 ontologies under DS. DSER imposes the so-called {\em Variable Typing}\footnote{\url{https://www.w3.org/TR/2013/REC-sparql11-entailment-20130321/}} constraint that disallows the use of the same variable in a different type of position and makes metaquerying impossible. In the relational calculus notation above it becomes clear that this query has a higher-order aspect to it, as a variable occurs in the position of a predicate.

To remedy the limitation of metamodeling, Higher-Order Semantics (HOS) was introduced in \cite{lenzerini2015higher} for OWL 2 QL ontologies. The interpretation structure of HOS follows the Hilog-style semantics of \cite{chen1993hilog}, which allows the elements in the domain to have polymorphic characteristics. Data complexity of HOS stays in \textbf{AC}$^{0}$ for answering unions of conjunctive queries, as shown in \cite{lenzerini2016higher}. In \cite{lenzerini2020metaquerying}, HOS is referred to as Meta-modeling Semantics (MS), which is the terminology that we will adopt in this paper. And to remedy the limitation of metaquerying, the Meta-modeling Semantics Entailment Regime (MSER) was proposed in \cite{cima2017sparql}, which does allow meta-modeling and meta-querying using SPARQL over OWL 2 QL.

In \cite{cima2017sparql} a reduction from query-answering over OWL 2 QL to Datalog queries is provided, and experimental results using two Datalog engines, LogicBlox, see \cite{aref2015design}, and RDFox, see \cite{nenov2015RDFox}, are reported. Datalog (see for instance \cite{ceri2012logic}) is an attractive target language, as it has received increased academic interest with renewed tool support over the last few decades, see for instance \cite{huang2011datalog,de2012datalog}. From logic programming (cf.\  \cite{lloyd2012foundations}), Datalog emerged as an eminent language. Datalog enjoys tractable reasoning with respect to the input data size (given a fixed program).  The key aspect of Datalog is to be utilized as an expressive querying language for relational data. As a result, Datalog has advanced into a ﬁrst-class formalism with eﬃcient implementations. Moreover, Datalog has also proven particularly relevant for Semantic Web applications such as ontological modeling and reasoning, as rules in Datalog can represent clauses in the fragment of function-free Horn First-Order Logic.

In this work, we address the practical effectiveness of the Datalog reduction of \cite{cima2017sparql}. In \cite{cima2017sparql} a preliminary performance analysis was reported, using the tools LogicBlox and RDFox on queries over LUBM ontologies. In \cite{lenzerini2020metaquerying}, meta-modeling and meta-querying in OWL 2 QL under MSER was evaluated as well using the larger MODEUS ontologies, which also contain more meta-modeling. However, that work did not evaluate the Datalog characterization of \cite{cima2017sparql}. In our work, we consider on the one hand a larger spectrum of tools that support Datalog (but stem from a variety of foundations and techniques), and on the other hand consider queries over both LUBM and MODEUS ontologies, expanding \cite{qureshi2021evaluation}. It turns out that other tools perform very well, too, especially in the meta-modeling heavy MODEUS scenarios, which were not considered in \cite{cima2017sparql}. So there are tools that allow for practical query answering also on sizable ontologies.

\subparagraph{\textbf{Contributions.}} This work builds upon \cite{cima2017sparql} and \cite{qureshi2021evaluation} and extends these by a detailed study of metamodeling and metaquerying in OWL 2 QL, along with the experiments and results. The main contributions of this work can be summarised as follows:

\begin{itemize}
\item We have extended the performance evaluation of the Datalog-based MSER approach discussed in \cite{qureshi2021evaluation} to three more back-end tools, the hybrid reasoning tools DLVHex and HexLite, and the lazy-grounding ASP solver Alpha, totalling nine back-end tools.
\item In \cite{qureshi2021evaluation} no time and memory limit was set. In this work, we consider two different limits, a \emph{Tight Resource Limit} and a \emph{More Generous Resource Limit}.
\item In the evaluation, we considered both meta-modeling ontologies and meta-queries. Furthermore, we have also considered a non-metamodeling ontology and standard query setting. The latter serves as a baseline, in order to understand the feasibility of MSER also in a settings in which it is equivalent to DSER.
    \item We give a detailed discussion on the performance and the evaluation results of the tools under the two limits.
\end{itemize}

\subparagraph{\textbf{Outline.}} The remainder of this paper is organized as follows. Section 2 introduces the background notions like OWL 2 QL, Meta-modeling Semantics (MS), SPARQL, and Datalog. Section 3 gives an overview of the rewriting presented in \cite{cima2017sparql,qureshi2021evaluation}. Section 4 illustrates the series of experiments we have carried out using the LUBM and MODEUS ontologies and presents a detailed discussion of the findings. Section 5 compares our approach with other approaches to metamodeling in the literature. Section 6 concludes the paper with a brief discussion and pointing out some future work.

\section{Preliminaries}
In this section, we recall the preliminary notions that are needed for the rest of the paper, in particular OWL 2 QL under MS, SPARQL, and Datalog.

\subsection{OWL 2 QL}
This section reviews the syntax of the ontology language OWL 2 QL and the \textit{Metamodeling Semantics} (MS) for OWL 2 QL, a profile of OWL 2. In general, OWL 2 QL is an easy-to-implement sub-language of OWL 2, obtained by decreasing the expressive power of OWL 2 DL, matching \textit{DL-Lite}$_{R}$, allowing for tractable reasoning and querying under direct semantics in AC$^{0}$ (data complexity). In this work we did not deal with data properties and literals and that is why did not include them in language definitions. We refer to \cite{lenzerini2021metamodeling} for a complete language definition.

\subparagraph{\textbf{Syntax.}} We start by recalling some basic elements used for representing knowledge in ontologies: \textit{individuals} ($i$), objects of a domain of discourse, \textit{concepts} or \textit{classes} ($c$), sets of individuals with common properties, and \textit{roles} or \textit{object properties} ($p$), relations that link individuals. Given the vocabulary $V$ of an OWL 2 QL ontology $V= (V_{n}, V_{c}, V_{p}, 
V_{i})$, where $V_{n}$ consists of Internationalized Resource Identifiers (IRIs), which are used for denoting classes, object properties, and individuals, grouped into sets $V_c, V_p, V_i$, respectively. $V_{c}$ (resp., $V_{p}$) includes the special symbols $\top_{c}$ (resp., $\bot_{c}$) and $\top_{p}$ (resp., $\bot_{p}$). We denote $Exp(V)$ the set of all well-formed expressions over $V$ written as $Exp(V) = V_{n} \cup Exp_{c}(V) \cup Exp_{p}(V) \cup V_{i}$, where
\begin{itemize}
    \item $Exp_{c}(V)$ = $V_{c} \cup \{\exists e_{1}.e_{2} \mid e_{1} \in Exp_{p}(V),e_{2} \in V_{C}\}$ are class expressions.
    \item $Exp_{p}(V)$ = $V_{p} \cup \{e^{-} \mid e \in V_{p}\}$ are object property expressions, containing either object properties or their inverses.
    \item the set of individual expressions coincides with $V_{i}$.
\end{itemize}

An OWL 2 QL \emph{ontology} is a finite set of logical axioms over $V$. The axioms allowed in an OWL 2 QL ontology have one of the form listed in the Table~\ref{axiom-table}, where (with optional subscripts) $\mathrm{\boldsymbol{ca, ra, i}}$ denote an atomic class, an atomic object property, an individual expression; $\mathrm{\boldsymbol{b}}$ denotes a class expression; $\mathrm{\boldsymbol{c}}$ and $\mathrm{\boldsymbol{p}}$ denote a class and an object property. Moreover, $\mathrm{\boldsymbol{re\!f}}$ and $\mathrm{\boldsymbol{irre\!f}}$ are keywords for declaring object properties to be reflexive and irreflexive, respectively. The Table~\ref{axiom-table} is divided into TBox axioms (further divided into positive TBox axioms and negative TBox axioms) and ABox axioms. 

\begin{table}[h!]
\caption{OWL 2 QL axioms}
\centering
\label{axiom-table}
\begin{tabular}{|ll|}
\noalign{\hrule height 0.5pt}
\multicolumn{1}{|l|}{\textbf{TBox (positive axioms)}}    & \textbf{TBox (negative axioms)}                        \\ \noalign{\hrule height 0.5pt}
\multicolumn{1}{|l|}{$\mathrm{\boldsymbol{b \sqsubseteq_{c} c}}\>\>(class\> inclusion)$}             & $\mathrm{\boldsymbol{b_{1} \sqsubseteq_{c} \neg b_{2}}}\>\>(class\> disjointness)$ \\ \noalign{\hrule height 0.5pt}
\multicolumn{1}{|l|}{$\mathrm{\boldsymbol{p_{1} \sqsubseteq_{p} p_{2}}}\>\>(object\> property\> inclusion)$} & $\mathrm{\boldsymbol{p_{1} \sqsubseteq_{p} \neg p_{2}}}\>\>(object\> property\> disjointness)$ \\ \noalign{\hrule height 0.5pt}
\multicolumn{1}{|l|}{$\mathrm{\boldsymbol{re\!f(p)}}\>\>(reflexive\> object\> property)$}     & $\mathrm{\boldsymbol{irre\!f(p)}}\>\>(irreflexive\> object\> property)$       \\ \noalign{\hrule height 0.5pt}
\multicolumn{2}{|c|}{\textbf{ABox axioms}}                                                                        \\ \noalign{\hrule height 0.5pt}
\multicolumn{1}{|l|}{$\mathrm{\boldsymbol{ca(i)}}\>\>(class\> assertion)$}     & $\mathrm{\boldsymbol{ra(i_{1},i_{2})}} \>\>(object\> property\> assertion)$                     \\ \noalign{\hrule height 0.5pt}
\end{tabular}
\end{table}

In Table~\ref{axiom-table} (i) the positive TBox axioms are the inclusion axioms, namely class inclusions, object property inclusions, plus the reflexivity declarations, (ii) negative TBox axioms are the disjoint axioms, namely class disjointness, object property disjointness, plus irreflexivity declarations, and  (iii) ABox axioms are membership assertion axioms, namely class membership assertion, object property membership assertion and different individual declarations. For simplicity, in this paper, we omit to deal with OWL 2 QL axioms that can be expressed by appropriate combinations of the axioms specified in Table~\ref{axiom-table}. Examples of such axioms are: equivalences, disjointness of more than two entities, or object property symmetry. Lastly, we assume that the symbols $\mathrm{\boldsymbol{ca,ra}}$ appearing in the ABox axioms of the ontology such as $\mathrm{\boldsymbol{ca(i)}}$ and $\mathrm{\boldsymbol{ra(i_{1},i_{2})}}$, also appear in the TBox of the same ontology without loss of generality (by having axioms like $\mathrm{\boldsymbol{ca \sqsubseteq_{c} \top_{c}, ra \sqsubseteq_{p} \top_{p}}}$.

Now, we return to the {\em punning} feature (the same entity can appear on the position of different types) in OWL 2 QL discussed in the Introduction. Every position in the axioms refers to a certain type, e.g.\ $\mathrm{\boldsymbol{c_{1} \sqsubseteq_{c} c_{2}}}$ is the {\em SubClassOf} axiom, where both positions are occupied by class expressions, therefore both are {\em class} positions. Similarly $\mathrm{\boldsymbol{ca(i)}}$ is a {\em ClassAssertion} axiom where first position in the axiom is a {\em class} position and the second one is an {\em individual} position. If we look at Example~\ref{ex:goldeneagle}, the entity \textbf{GoldenEagle} appears as the first argument of a {\em SubClassOf} axiom and the second argument of a {\em ClassAssertion} axiom, thus the former position refers to a {\em class} position and the latter refers to an {\em individual} position.

\subsection{Meta-modeling Semantics}

The Meta-modeling Semantics (MS) is based on the idea that every entity in $V$ may simultaneously have more than one type, so it can be a class, or an individual, or data property, or an object property or a data type. To formalise this idea, the Meta-modeling Semantics has been defined for OWL 2 QL. We refer to \cite{lenzerini2021metamodeling} for a comprehensive definition.

The meta-modeling semantics for $\mathcal{O}$ over $V$ is based on the notion of interpretation, constituted by a tuple $\mathcal{I} = \langle \Delta, \cdot^I, \cdot^C, \cdot^P, \cdot^\mathcal{I}\rangle$, where ($\mathcal{P}(S)$ denotes the power set of $S$)
\begin{itemize}
    \item $\Delta$ is the union of the two non-empty disjoint sets: $\Delta = \Delta^{o} \cup \Delta^{v}$, where $\Delta^o$ is the object domain, and $\Delta^{v}$ is the value domain. Each element in the world is either object or a value;
    \item $\cdot^I : \Delta^o \to \{True, False\}$ is a total function for each object $o \in \Delta^o$, which indicates whether $o$ is an individual; if $\cdot^C, \cdot^P, \cdot^D, \cdot^T$ are undefined for some $o$, then we require $o^{I} = True$, also in a few other cases such as if $o$ is in the range of $\cdot^C$;
    \item $\cdot^C : \Delta^o \to \mathcal{P}(\Delta^o)$ is partial and can assign the extension of a class;
    \item $\cdot^P : \Delta^o \to \mathcal{P}(\Delta^o \times \Delta^o)$ is partial and can assign the extension of an object property;
    \item $^{.\mathcal{I}}$ is a function that maps every expression in $Exp(V)$ to $\Delta^o$ and every literal to $\Delta^{v}$.
\end{itemize}

There are a couple of constraints on interpretations, for which we refer to Table 3 in \cite{lenzerini2021metamodeling}. Note that we follow  \cite{lenzerini2021metamodeling} also with naming the interpretation $\mathcal{I}$ while using the same symbol for one of its component functions, which can be distinguished by the fact that the latter is used only in superscripts. 

This allows for a single object $o$ to be simultaneously interpreted as an individual via $^.{}^I$, a class via $^.{}^C$ and an object property via $^.{}^P$. For instance, for Example~\ref{ex:goldeneagle}, $\cdot^C,\>\cdot^I$ would be defined for \textit{GoldenEagle}$^{\mathcal{I}}$, while $\cdot^{P}$ would be undefined for it.

The semantics of a logical axiom $\alpha$ is defined in accordance with the notion of axiom satisfaction for the MS interpretation $\mathcal{I}$. The language is specified in Table 3.B in \cite{lenzerini2021metamodeling}. $\mathcal{I}$ is a model of ontology $\mathcal{O}$ if it satisfies all axioms of $\mathcal{O}$. Finally, an axiom $\alpha$ is said to be logically implied by $\mathcal{O}$, denoted as $\mathcal{O} \models  \alpha $, if it is satisfied by every model of $\mathcal{O}$.

\subsection{The query language}
As query language, we consider conjunctive queries expressed using SPARQL. Conjunctive queries are based on the notion of query atoms, which is based on the notion of query terms (may involve variables). Given an ontology over $V$ and set of variables $\mathcal{V}$ disjoint from $V$, a query atom (or atom) has the same form as axioms with the difference that query terms (expressions of the set $Exp(V \cup \mathcal{V})$) may occur in it. 

A conjunctive \textit{SPARQL} query \textit{q} is of the form
\[
SELECT\ ?v_1\ \cdots\ ?v_n\  WHERE\ \{B\}
\]
where  $\{?v_{1},...,?v_{n}\} \subseteq \mathcal{V}$ and $B$ called body of conjunctive query $q$, denoted as $body(q)$, consisting of a conjunction of triples that are to be read as atoms over terms $Exp(V \cup \mathcal{V})$, and $\{?v_{1},...,?v_{n}\}$ has to be a subset of the variables occurring in $B$. 
Given an ontology $\mathcal{O}$ and a conjunctive SPARQL query $q$, the answers to $q$ are the tuples $\langle ?v_1\sigma, \ldots, ?v_n\sigma\rangle$, where $\sigma$ is a substitution that maps all variables in $q$ to elements of $V$ such that $\mathcal{O} \models B\sigma$. 

Note that SPARQL usually uses RDF triple syntax, which we will also write as atoms.
In the following we will sometimes use the notation $q(v_1, \ldots, v_n) \gets B'$ to denote a query of the form above, where $B'$ corresponds to $B$ but is written as a conjunction of atoms.

We would like to distinguish between standard queries and meta-queries. 
A meta-query is a query consisting of meta-predicates $p$ and meta-variables $v$, where $p$ can have other predicates as their arguments and $v$ can appear in predicate positions. For instance, if $c$ is a class name and $x,y$ are variables, query
\[
q(x)\ \gets c(x) \land \ x(y).
\]
  will return all classes that are themselves members of another class.

The simplest form of meta-query is a query where variables appear in class or property positions also known as \emph{second-order queries}. More interesting forms of meta-queries allow one to extract complex patterns from the ontology, by allowing variables to appear simultaneously in individual object and class or property positions. We will refer to non-meta-queries as standard queries. 
  
As for the semantics of query $\mathcal{Q}$ under MS, we rely on the definitions of \cite{lenzerini2021metamodeling} and do not provide any formal details here. Essentially, given a query $\mathcal{Q}$ expressed over $\mathcal{O}$,  $\mathcal{O} \models_{MS} \mathcal{Q}\sigma$ holds if $\mathcal{Q}\sigma$ is true in every MS model of $\mathcal{O}$ for a substitution $\sigma$ for the variables in $\mathcal{Q}$.

\subsection{Datalog}
Datalog is a declarative query language rooted in logic programming. Datalog has similar expressive power as SQL, whose more recent versions also allow to express recursive queries over a relational database, previously a distinguishing feature of Datalog.

\subparagraph{\textbf{Syntax.}}\mbox{}\\
\mbox{}\\
 A Datalog program consists of a finite set of facts and rules, where facts are seen as the assertions about the world, i.e. {\em “Harry is an instance of GoldenEagle”} and rules are sentences that allow for deducing facts from other facts, i.e. “{\em If X is an instance-Of Y, and Y has a parent Z, then X is an instance-Of Z”}.  Generally, rules contain variables like X, Y, or Z, and when combining both facts and rules, they form a knowledge base.

A Datalog program is a finite set of rules $r$, where both facts and rules are represented as Horn clauses of the form \[h \gets b_{1},\ldots,b_{n}.\] In $r$, $h$ is the rule head and $b_{1},\ldots,b_{n}$ is the rule body. Each of $b_{i}$ and $h$ are atoms of the form $p(t_{1},\ldots,t_{n})$, where $p$ is a predicate symbol and $t_{i}$ are \textit{terms} that can be \textit{constants} or \textit{variables}; $n$ is the arity of $p$. We say that $r$ is a fact if it has an empty body and omit the symbol $\gets$ in that case, for example \[instc(GoldenEagle,Harry).\] The body can be thought of as a conjunction, for example in \[instc(Z,X) \gets instc(X,Y),isacCC(Y,Z).\]
In these examples, \textit{GoldenEagle, Harry} are the constants, \textit{X, Y, Z} are the variables and \textit{instc, isacCC} are the predicates. A literal, fact, or rule, which does not contain any variables, is called ground. 

Rules need to satisfy safety in order to be domain-independent and to guarantee that the grounding of the Datalog program is finite, so each variable occurring in the rule head must also occur in the rule body and facts may not contain variables.

\subparagraph{\textbf{Semantics}}\mbox{}\\
\mbox{}\\
For a Datalog program $P$, the Herbrand Base $HB(P)$ is the set of all atoms of the form $p(c_{1},...,c_{n})$, where $p$ and $c_{i}$ are predicates and constants in $P$. An interpretation $I \subseteq HB(P)$ consists of those atoms that are true in $I$. A positive ground literal $l$ (resp. $\neg l$) is satisfied by $I$ if $l \in I$ (resp. $l \notin I$). A rule $r$ is satisfied if $h \in I$ whenever $\{b_{1},\ldots,b_{n}\} \subseteq I$, and an interpretation $I$ is a model of a program $P$ if all of its rules are satisfied. The semantics of $P$ is given by the subset-minimal model of $P$, which is guaranteed to exist and to be unique. 

A list of ground atoms $a_{1},...,a_{n}$, interpreted as conjunction of atoms are said to be logical consequences of $P$ denoted as $P \models a_{1},...,a_{n}$ if each $a_{i}$ is satisfied in all models of $P$, which means that each $a_{i}$ is satisfied in the subset-minimal model.

A query for a Datalog program D is a conjunction of atoms written as:
\[q_1, \ldots, q_n?\] It consists of atoms as in Datalog rules, and the answers with respect to a program $P$ are those substitutions $\sigma$ (mapping variables in the query to constants) such that $\{q_1\sigma, \ldots, q_n\sigma\} \subseteq M$ for the  subset-minimal model $M$ of $P$. Equivalently, the answers are those $\sigma$ such that $P \models q_1\sigma, \ldots, q_n\sigma$.

While there are dedicated Datalog-only systems, Datalog is also the core of several richer languages and systems that support them. This is the main motivation of our work, in which we examine how various tools that support Datalog query answering perform on Datalog queries stemming from meta-querying over OWL 2 QL ontologies.

\section{Overview of the Approach}\label{sec:overview}
In this section, we recall query answering under the Meta-modeling Semantics Entailment Regime (MSER) from \cite{cima2017sparql}. This technique reduces  SPARQL query answering over OWL 2 QL ontologies to Datalog query answering.  The main idea of this approach is to define (i)  a translation function $\tau$ mapping OWL 2 QL axioms to Datalog facts and (ii)  a fixed rule base $\mathcal{R}^{ql}$ that captures inferences in OWL 2 QL reasoning.  Note that, for the sake of simplicity, we follow \cite{cima2017sparql} and do not consider ontologies containing data properties for easier presentation.

The reduction employs a number of predicates, which are used to encode the basic axioms available in OWL 2 QL. This includes both axioms that are explicitly represented in the ontology (these will be added to the Datalog program as facts resulting from the mapping $\tau$) and axioms that logically follow (these will be derivable by the fixed rules $\mathcal{R}^{ql}$). In a sense, this representation is closer to a meta-programming representation than other Datalog embeddings that translate each axiom to a rule.

The function $\tau$ is used to encode the OWL 2 QL assertions $\alpha$ as facts. For a given ontology $\mathcal{O}$, we will denote the set of facts obtained by applying $\tau$ to all of its axiom as $\mathcal{P}^{ql}_{\mathcal{O}}$; it will be composed of two portions $\mathcal{P}^{ql,\mathcal{T}}_{\mathcal{O}}$ and $\mathcal{P}^{ql,\mathcal{A}}_{\mathcal{O}}$, as indicated in Table~\ref{tau-function}.

\begin{table}[h!]
\centering
\caption{$\tau$ Function}
\label{tau-function}
\begin{tabular}{|c|l|l|ll|}
\noalign{\hrule height 0.5pt}
$\mathcal{P}^{ql}_{\mathcal{O}}$                               & \multicolumn{1}{c|}{$\alpha$}                  & \multicolumn{1}{c|}{$\tau$($\alpha$)} & \multicolumn{1}{c|}{$\alpha$}                                         & \multicolumn{1}{c|}{$\tau$($\alpha$)} \\ \noalign{\hrule height 0.5pt}
\multirow{12}{*}{$\mathcal{P}^{ql,\mathcal{T}}_{\mathcal{O}}$} & c1 $\sqsubseteq$ c2                            & isacCC(c1, c2)                        & \multicolumn{1}{l|}{r1 $\sqsubseteq \neg$ r2}                         & disjrRR(r1,r2)                        \\ \cline{2-5} 
                                                               & c1 $\sqsubseteq \exists$r2$^-$.c2              & isacCI(c1, r2, c2)                    & \multicolumn{1}{l|}{c1 $\sqsubseteq \neg$ c2}                         & disjcCC(c1,c2)                        \\ \cline{2-5} 
                                                               & $\exists$r1 $\sqsubseteq \exists$r2.c2         & isacRR(r1,r2,c2                       & \multicolumn{1}{l|}{c1 $\sqsubseteq \neg \exists$r2$^-$}              & disjcCI(c1,r2)                        \\ \cline{2-5} 
                                                               & $\exists$r1$^- \sqsubseteq$ c2                 & isacIC(r1,c2)                         & \multicolumn{1}{l|}{$\exists$r1$\sqsubseteq \neg$ c2}                 & disjcRC(r1,c2)                        \\ \cline{2-5} 
                                                               & $\exists$r1$^-$ $\sqsubseteq \exists$r2.c2     & isacIR(r1,r2,c2)                      & \multicolumn{1}{l|}{$\exists$r$_1$ $\sqsubseteq \neg \exists$r2}      & disjcRR(r1,r2)                        \\ \cline{2-5} 
                                                               & $\exists$r1$^-$ $\sqsubseteq \exists$r2$^-$.c2 & isacII(r1,r2,c2)                      & \multicolumn{1}{l|}{$\exists$r1 $\sqsubseteq \neg \exists$r2$^-$}     & disjcRI(r1,r2)                        \\ \cline{2-5} 
                                                               & r1 $\sqsubseteq$ r2                            & isarRR(r1,r2)                         & \multicolumn{1}{l|}{$\exists$r1$^- \sqsubseteq \neg$ c2}              & disjcIC(r1,c2)                        \\ \cline{2-5} 
                                                               & r1 $\sqsubseteq$ r2$^-$                        & isarRI(r1,r2)                         & \multicolumn{1}{l|}{$\exists$r1$^-$ $\sqsubseteq \neg \exists$r2}     & disjcIR(r1,r2)                        \\ \cline{2-5} 
                                                               & c1 $\sqsubseteq \exists$r2.c2                  & isacCR(c1,r2,c2)                      & \multicolumn{1}{l|}{$\exists$r1$^-$ $\sqsubseteq \neg \exists$r2$^-$} & disjcII(r1,r2)                        \\ \cline{2-5} 
                                                               & $\exists$r1$\sqsubseteq$ c2                    & isacRC(r1,c2)                         & \multicolumn{1}{l|}{r1 $\sqsubseteq \neg$ r2$^-$}                     & disjrRI(r1,r2)                        \\ \cline{2-5} 
                                                               & $\exists$r1 $\sqsubseteq \exists$r2$^-$.c2     & isacRI(r1,r2,c2)                      & \multicolumn{1}{l|}{irref(r)}                                         & irrefl(r)                             \\ \cline{2-5} 
                                                               & refl(r)                                        & refl(r)                               &                                                                       &                                       \\ \noalign{\hrule height 0.5pt}
\multirow{2}{*}{$\mathcal{P}^{ql,\mathcal{A}}_{\mathcal{O}}$}  & c(x)                                           & instc(c,x)                            & \multicolumn{1}{l|}{x $\neq$ y}                                       & diff(x,y)                             \\ \cline{2-5} 
                                                               & r(x, y)                                        & instr(r,x,y)                          &                                                                       &                                       \\ \noalign{\hrule height 0.5pt}
\end{tabular}
\end{table}

The fixed Datalog program $\mathcal{R}^{ql}$ can be viewed as an encoding of axiom saturation in OWL 2 QL. The full set of rules were provided by the authors of \cite{cima2017sparql} and reported in \cite{qureshi2021evaluation}. They can also be found in the online repository \url{https://doi.org/10.5281/zenodo.7286886}. We will consider one rule to illustrate the underlying ideas:

\begin{quote}
   \centering isacCR(c1,r2,c2) :- isacCC(c1,c3), isacCR(c3,r2,c2)
\end{quote}

The above rule encodes the following inference rule:
\begin{quote}
   \centering $\mathcal{O} \models$ c1 $\sqsubseteq$ c3, $\mathcal{O} \models$ c3 $\sqsubseteq \exists$r2.c2 $ \Rightarrow  \mathcal{O} \models$ c1 $\sqsubseteq \exists$r2.c2
\end{quote}

Finally, the translation can be extended in order to transform conjunctive SPARQL queries under MS over OWL 2 QL ontologies into a Datalog query. For example, consider the following query that retrieves all triples $\langle x,y,z\rangle$, where $x$ is a member of class $y$ that is a subclass of $z$: \\

SELECT$\>?x\> ?y\> ?z\>$WHERE $\>\{ \\
    \indent  \indent  \indent ?x \> rdf\!:\!type \> ?y .\\
    \indent  \indent  \indent ?y \> rdfs\!:\!SubClassOf \> ?z \\
     \indent \indent \}$\\

     This can be translated to a Datalog query
     
\begin{quote}
   \centering instc(X,Y), isacCC(Y,Z)?
\end{quote}

In general, these queries will be translated into a rule plus an atomic query to account for projections. The previous example will be translated to

\begin{quote}
  \centering q(X,Y,Z) $\gets$ instc(X,Y), isacCC(Y,Z).\\
  q(X,Y,Z)?
\end{quote}

\section{Experimental Setup}\label{exp}
In this section we describe the extended experiments that we have conducted, including the tools we used, the ontologies and queries we considered, and report on the outcomes. All material is available at \url{https://doi.org/10.5281/zenodo.7286886}.

\subsection{Tools}
We have implemented the translation of ontology axioms summarised in Table~\ref{tau-function} in Java. We should point out that this implementation is not optimised and serves as a proof of concept. For the Datalog back-end, we have tested six tools from \cite{qureshi2021evaluation} and add three additional tools in our experiments. Our decisions with respect to tool selection were based on problem statement analysis and preliminary performance assessments. These tools stem from different logic paradigms like Prolog, pure Datalog, Answer Set Programming and Hybrid-Knowledge Bases.  In the following, we briefly describe each of these tools. For more detailed descriptions we refer interested readers to \cite{qureshi2021evaluation}.

\subsection{Datalog}
A common practice across all these Datalog systems is the use of the Datalog language in the area of efficient query answering, formal reasoning and analysis, where tools such as \textbf{RDFox} and \textbf{LogicBlox} have shown great results. 

\subsubsection{RDFox}
\textit{RDFox} \cite{nenov2015RDFox} is an in-memory, scalable, centralised data engine for Resource Description Framework (RDF) data models. The tool supports the current standard querying language  SPARQL 1.1. It also allows for reasoning and representing knowledge in rules, supporting materialisation-based parallel Datalog reasoning. RDFox uses parallel reasoning algorithms to support Datalog reasoning over RDF data.

\subsubsection{LogicBlox}
\textit{LogicBlox} \cite{aref2015design} is another state-of-the-art Datalog engine to allow applications to automate and enhance their decision making via a single expressive declarative language. LogicBlox supports module mechanisms and uses non-trivial type systems to detect programming errors and aid query optimization \cite{de2008type}.

\subsection{Prolog}
Prolog is the classic logic programming language. Datalog was motivated by Prolog and can be viewed as a subset of it. While using Prolog naively on Datalog queries will lead to termination issues, techniques such as tabling alleviate this. Its application in various problem-solving areas has demonstrated its capabilities due to increasingly efficient implementations and ever richer environments such as \textbf{XSB}.

\subsubsection{XSB}
\textit{XSB} \cite{swift2012xsb} is a logic programming engine and language rooted in Prolog. It supports Prolog's standard functionality, and features a powerful technique called tabling, which significantly increases its applicability and is particularly relevant for Datalog query answering. Since it relies on a top-down technique, its internals are significantly different from RDFox and LogicBlox.

\subsection{Hybrid Approaches}
Hybrid approaches are interesting because they support the DL-based knowledge representation languages at the core of the semantic web, and at the same time, they overcome some shortcomings of DLs, where rules are better suited. The hybrid solutions range from {\em loosely coupled} formalisms, such as those in \cite{eiter2008combining,wang2004combining}, where the rule parts and DL parts of the knowledge base are treated like separate components that interact by querying each other, to fully integrated approaches, or {\em tightly coupled} formalisms such as \cite{bruijn2011embedding,motik2007faithful}, where the vocabulary and semantics of rule and DL parts are homogeneous. In this work we focus on the hybrid formalisms \textbf{NoHR}.

\subsubsection{NoHR}
\textit{NoHR} \cite{lopes2017nohr}, the Nova Hybrid Reasoner allows to query a combination of DL ontologies (OWL 2 EL or OWL 2 QL) and non-monotonic rules in a top down fashion. NoHR is the first hybrid reasoner of its kind for Protégé. We have included NoHR because we wanted to assess the overhead it produces with respect to XSB, as the agenda for our future work includes leveraging hybrid tools like NoHR for meta-modeling and meta-querying.

\subsection{Answer Set Programming (ASP)}
Answer Set Programming (ASP) \cite{breitr11a} is a declarative programming paradigm that applies non-monotonic reasoning. ASP falls into the category of declarative and logic programming, and can be viewed as an extension to Datalog, which handles features such as negation as failure, disjunction in a semantically clear way. Its main applications have been in solving combinatorial problems, but it has also been successfully used for reasoning in knowledge representation and databases. The success of ASP is partially due to efficient solver technology available for evaluation, for instance \textbf{Clingo}, \textbf{DLV2}, \textbf{Alpha}, \textbf{DLVHex}, and \textbf{HexLite}. Most state-of-the-art ASP systems follow the {\em ground and solve} approach, where an input program is turned into a corresponding variable-free (ground) program for which answer sets are then computed. A few ASP systems use a {\em lazy-grounding} technique \cite{weinzierl2019alpha,dao2012omiga}, where the main idea is to interleave grounding with solving and generate only ground rules necessary in each position of the search space, for instance {\em Alpha}.

\subsubsection{Clingo}
\textit{Clingo} \cite{gebser2010gringo} is probably the currently most widely used tool for Answer Set Programming. It follows the ground and solve approach, with subsystems \textit{gringo} and \textit{clasp} for grounding and solving, respectively.

\subsubsection{DLV2}
\textit{DLV2} \cite{alviano2017asp} is another ASP system with particular emphasis on Disjunctive Logic Programming and based on the declarative programming language datalog, known for being a convenient tool for knowledge representation. It is also based on the ground and solve approach.

\subsubsection{Alpha}
\textit{Alpha} \cite{weinzierl2017blending} is an ASP system that reads logic programs and computes the corresponding answer sets. In contrast to many other ASP systems, {\em Alpha} implements a lazy-grounding approach to overcome memory limits when working with considerable large input. Lazy grounding is a novel way to address the bottleneck in ground and solve approaches, by attempting to ground only what is needed. {\em Alpha} is not the quickest framework available, as it is intended as a testbed and prototype for lazy-grounding techniques. However, there are domains in which {\em Alpha} succeeds, where ground and solve systems fail completely.

\subsubsection{DLVHex}
\textit{DLVHex} \cite{eiter2006dlvhex}, a logic programming based reasoner, computes the models of so-called HEX-programs. The HEX-programs extend the Answer Set Programming (ASP) paradigm using non-monotonic logic programs with bidirectional access to external sources. Intuitively, the logic program sends information given by constants or predicate extensions to the external source, giving back output values imported into the program. It uses the {\em conflict-driven} technique for the evaluation of a program, to find an assignment that satisfies all {\em nogoods} (set of literals which must not be true simultaneously, see \cite{gebser2012conflict}). The \textit{DLVHex} system in its current version supports all features defined in the ASP standard, including function symbols, choice rules, conditional literals, aggregates, and weak constraints.

\subsubsection{HexLite}
The \textit{HexLite} \cite{schuller2019hexlite} solver is a lightweight alternative to \textit{DLVHex} with a restricted set of external computations. The intention is to provide a lightweight system for an easy start with HEX. The main aim of \textit{HexLite} is to achieve efficiency and simplicity, both in the implementation and installation of the system. It uses the \textit{pragmatic-hex-fragment} that permits to separate external computations into two parts: first, deal with those that can be evaluated during the instantiation, and second, deal with those that can be evaluated during the search. \textit{HexLite} is implemented in Python and uses the \textit{Clingo} Python API as a back-end for ASP grounding and search.

\subsection{Computational Settings}
The versions of the evaluated systems are: RDFox 5.1.0, LogicBlox 4.34.0, XSB 4.0, NoHR  3.0, Clingo  5.2.2, DLV2 2.0, Alpha  0.5.0, DLVHex 2.5.0, HexLite 1.4.1 (and the hexlite-owl-api plugin version is 1.2).

We observe that the tools slightly differ in the Datalog syntax they require and thus needed minor adjustments. For instance, LogicBlox uses \verb|<-| instead of \verb|:-|, and also variables are denoted in different ways in the various input languages. RDFox and XSB needed some major adjustments, which are described in detail in \cite{qureshi2021evaluation}. All other tools received the same input, except for variations due to language syntax.

\subsubsection{Benchmark}
Instead of creating a new benchmark from scratch, we decided to build our experiments on top of two datasets.

{\em Lehigh University Benchmark} (\textbf{LUBM}) was developed to facilitate the evaluation of Semantic Web reasoners in a standard and systematic way, with respect to extensional queries over synthetic OWL data scalable to arbitrary sizes, see \cite{guo2005lubm}. It describes a university domain with departments, courses, students, and faculty information. Although LUBM is a fairly simple ontology and does not contain meta-axioms, it captures existentially quantified knowledge that we find interesting to evaluate with different tools under MSER. Please note that MSER is not restricted to metamodeling features in ontology only, and we used LUBM to show the performance of different tools with non-metamodeling ontology. LUBM comes with a predefined data generator to generate random sizes of $\mathcal{A}$, which can be used to test the system ability of handling data of varying sizes. We have used LUBM(1) consisting of 43 classes, 32 properties (including 25 object properties and 7 datatype properties) and 10334 axioms and LUBM(9) consisting of 43 classes, 32 properties (including 25 object properties and 7 datatype properties) and 79501 axioms. The main difference between LUBM(1) and LUBM(9) is the size of the ABox. It uses OWL Lite language features including \textit{inverseOf}, \textit{someValuesFrom} restrictions, and \textit{intersectionOf}.

{\em Making Open Data Effectively USable} (\textbf{MODEUS}) contains four ontologies describing the \textit{Italian Public Debt} domain, designed in a joint project with the Italian Ministry of Economy and Finance, with information like financial liability or financial assets to any given contracts, see \cite{lenzerini2020metaquerying}. The MODEUS ontologies are relatively new and more complex than LUBM as it contains not only the meta-axioms but also complex patterns equipped with a spectrum of metamodeling features. We want to clarify that the MODEUS ontologies are not standard benchmark ontologies; we are using these ontologies as a benchmark for this work. The base ontology was progresively extended to obtain three larger ontologies. The first ontology \textbf{MEF$\_$00} contains 92 classes, 40 meta-classes, 11 properties, about 27000 individuals, 479 TBox axioms and about 137000 axioms. The second ontology \textbf{MEF$\_$01} obtained from \textbf{MEF$\_$00} by adding only 7000 fresh individuals and 35000 ABox axioms, which makes about 33000 individuals, about 172000 ABox axioms and the rest is same in that it contains 92 classes, 40 meta-classes and 11 properties. The third ontology \textbf{MEF$\_$02} obtained from \textbf{MEF$\_$00} by adding new classes, subset classes and disjoints, which makes 97 classes, 43 meta-classes, 11 properties, about 26000 individuals, 542 TBox axioms and about 137000 ABox axioms. The fourth ontology \textbf{MEF$\_$03} obtained from \textbf{MEF$\_$02} by adding 4000 individuals and 20000 ABox axioms, which give us 98 classes, 43 meta-classes, 11 properties, about 30000 individuals, 542 TBox axioms and about 157000 ABox axioms. MODEUS ontologies are meta-modeling ontologies with meta-classes and meta-properties.

\subsection{Benchmark Queries}
For the queries, we used the queries given by these datasets and some additional queries (or meta-queries) for LUBM. We referred the queries to their proper sites.

The LUBM dataset comes with standard 14 queries\footnote{http://swat.cse.lehigh.edu/projects/lubm/queries-sparql.txt} that contain queries with variables only
in place of individuals. These queries represent a variety of properties and several performance metrics like low selectivity vs high selectivity, implicit relationships vs explicit relationships, small input vs large input, etc. We have also considered the queries \textit{mq1}, \textit{mq4}, \textit{mq5}, and \textit{mq10} from \cite{kontchakov2014answering} and called them meta-queries as they comprise variables in predicate position i.e over class and property names. We have also considered two special case queries and called them special case meta-queries\footnote{The reason for this naming convention is, these queries are specifically created to check the meta-modeling case in LUBM-ext ontology.} \textit{sq1}, and \textit{sq2} from \cite{cima2017sparql} to test the meta-modeling features and identify the new challenges introduced by the additional expressivity over the ABox queries.

In special-case meta-queries, we check the impact of meta-classes and \texttt{DISJOINTWITH} in a query. For this, like in \cite{cima2017sparql}, we have introduced in both LUBM(1) and (9) a new class named \textit{TypeOfProfessor} and make \textit{FullProfessor}, \textit{AssociateProfessor} and \textit{AssistantProfessor} an instance of this new class, which makes \textit{TypeOfProfessor} a meta-class and also we define \textit{FullProfessor}, \textit{AssociateProfessor} and \textit{AssistantProfessor} to be disjoint from each other. Then, in \textit{sq1} we asked for all those $y$ and $z$, where $y$ is a professor, $z$ is a type of professor and $y$ is an instance of $z$, violating the typing constraint. In \textit{sq2}, we asked for instances of type of professor that are disjoint to see the impact of negative term for the evaluation of the query. We referred the extension of LUBM ontologies with additional expressivity as LUBM-ext, LUBM(1)-ext and LUBM(9)-ext. 

The MODEUS dataset comes with 9 queries\footnote{http://www.modeus.uniroma1.it/modeus/node/6}. These are all meta-queries, containing variables referring to more than one type, these queries traverse over several layers of \emph{instanceOf} relation. These queries represent the chain of \emph{instanceOf} relation, where some chains are expressed as \emph{classAssertion}, start with a specific entity, start with \emph{subclass}, asking for meta-property and some constituted by joining more than one chains of \emph{instanceOf}.

The queries \textit{mq1}, \textit{mq4}, \textit{mq5}, and \textit{mq10} are simpler forms of meta-queries that do not span over different levels of ontologies, as the variables appearing as class and property names in these queries do not appear in other types of positions. The queries \textit{sq1}, \textit{sq2} and  the 9 MODEUS queries are pure forms of the meta-queries that span over multiple level of ontologies.

\subsubsection{Test Environment}
We have done the test on a Linux batch server, running Ubuntu 20.04.3 LTS (GNU/Linux 5.4.0-88-generic x86\_64) on one AMD EPYC 7601 (32-Core CPU), 2.2GHz, Turbo max. 3.2GHz. The machine is equipped with 515GB RAM and a 4TB hard disk. Java applications used OpenJDK 11.0.11 with a maximum heap size of 25GB.

To perform different settings and measurements on the system resources we have used {\em Pyrunlim}\footnote{\url{https://github.com/alviano/python/tree/master/pyrunlim}} and {\em BenchExec}\footnote{\url{https://github.com/sosy-lab/benchexec}}. The reason for using two different tools for resource limitation is due to \textit{LogicBlox}, because it creates spawned processes, which causes resource consumption to be partially unaccounted for with \textit{Pyrunlim}. So, to limit the spawned processes on Linux, we have used the cgroups features of resource limitation and measurement provided by \textit{BenchExec}. Since we ran the \textit{LogicBlox} experiments last and encountered this issue, we did not re-run the other experiments with \textit{BenchExec}, as there should not be any dependence on the measurement tool. We manually verified on selected examples that  \textit{Pyrunlim} and \textit{BenchExec} yield the same results up to measurement precision.

\subsubsection{Performance Breakdown}
During the course of the evaluation of Datalog-based MSER we have used the two different resource limitations (on \textbf{RAM} and \textbf{Time}) as the benchmark setting on our data sets to examine the behavior of different tools with tight or ample resources. Thus, we can provide additional insight on the specific gains obtained by integrating various settings. We report the time of computing the answers to each query and include loading time.  For simplicity, we have not included queries that contain the data properties in our experiments.

If not otherwise indicated, in the first experiment, each benchmark had 15 minutes and 8GB of memory on a Linux cluster. We report the time of computing the answer sets for each query. In the second experiment, we increased the time to 30 minutes and 128GB of memory. 

\section{Experimental Results}

We now report the experimental results under the two resource bounds, starting with the tighter limits, followed by the more generous ones.

\subsection{Experiment 1 -- Tight Resource Limits}

We next report the results on different datasets under the first experiment setting,  first  LUBM  with standard queries\footnote{As one of the reviewer pointed out--this setting is interesting that it allows to compare the cost of evaluating meta-queries w.r.t standard queries using different Datalog engines, over an ontology that does not contain any meta-predicate.}, followed by meta-queries over LUBM, then moving towards MODEUS with meta-queries.

In the first batch of experiments for each tool, we have limited the RAM usage to 8GB and allowed for a maximum of 15 minutes to answer the query. In Figure 1 we have reported the results for different ABox sizes of the LUBM ontologies. All times reported in these tables are in seconds and include loading the Datalog program including facts and rules and answering the query. The best performance for each query is highlighted in boldface. The value \textit{OOT} and \textit{OOM} refers to \textit{Out-Of-Time} and \textit{Out-Of-Memory}.

\subsubsection{LUBM with standard-queries and simple meta-queries}
We can observe that performance is satisfactory for LUBM(1). Of these 14 queries, q2,q7, and q9 are the most interesting ones since they contain many atoms affecting running time. In these queries, we observe an increase in running time for some tools. We can see, however, that LogicBlox, NoHR, HexLite, and Alpha generally introduce significant overhead, which we did not really expect on this scale. 

\begin{figure}[!h]
  \centering
  \subfloat[LUBM(1)]{\includegraphics[width=6.2cm]{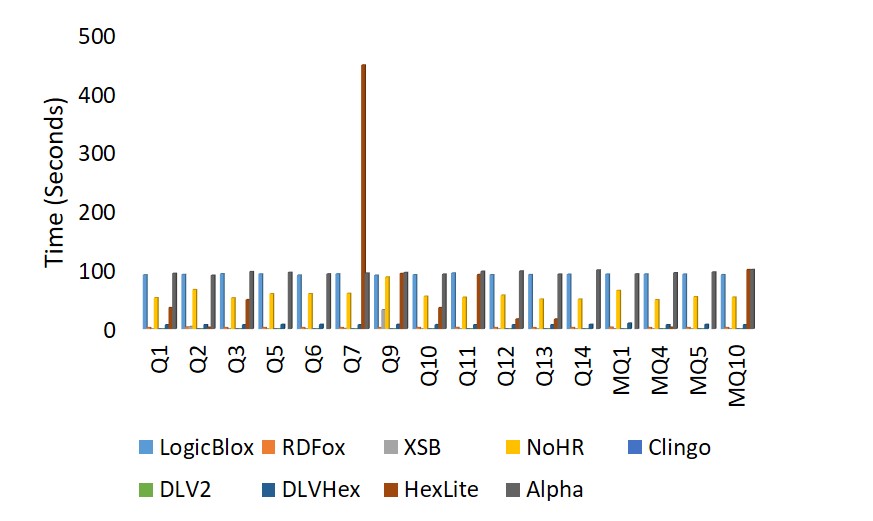}\label{fig:f1}}
  \hfill
  \subfloat[LUBM(9)]{\includegraphics[width=6.2cm]{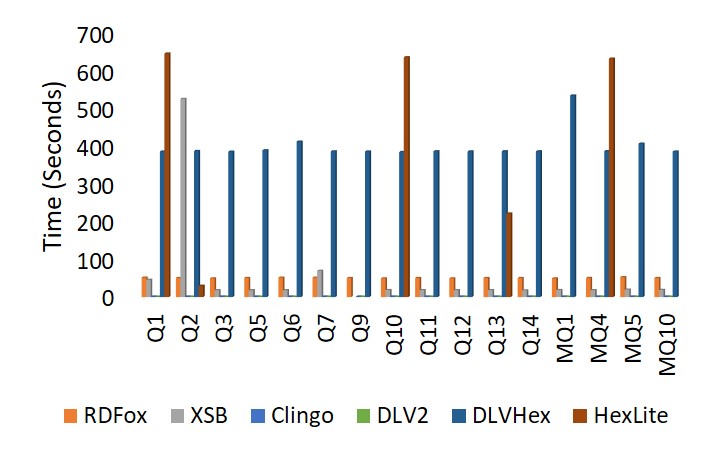}\label{fig:f2}}
  \caption{Experiment 1 LUBM with Standard Queries and Simple Meta-Queries}
  \label{E1L}
\end{figure}

It is not fully clear to us yet what causes this;
the ontology part is empty in the hybrid system NoHR, so we would have expected NoHR performance to be similar to XSB. It might be that the tabling strategy is to blame in LUBM(1) and in LUBM(9), we have observed that time and tabling strategy is directly proportional to each other, but this is just a conjecture. There is one instance in which XSB is quite slow (q9 on LUBM(1) and LUBM(9)), but performance on this query is generally bad and a different indexing strategy might be the cause for this behaviour. For HexLite, we would have expected more or less the same performance to DLVHex/Clingo as the queries were simple, there is not much external computation involved, and it uses Clingo for searching and grounding. It might be possible that the external OWL-API plugin that we used with HEXLite caused an overhead for HEXLite. As for Alpha, it is not an optimised system, but we were expecting better performance than other ASP tools. Furthermore, Alpha performance degraded in LUBM(9) when the input data size increases and we could not observe benefits of lazy grounding in our experiments.

Across all queries of  LUBM(1), LogicBlox exhibits very stable performance with roughly the same execution time for all queries, which is quite remarkable. But it lost this performance on LUBM(9), where the ontology size has affected its performance. Also, DLV2 and Clingo have very regular performance, but the time is affected slightly by the size of the dataset. RDFox is also very regular across different queries but is quite a bit affected by the size of the dataset. DLVHex, HexLite, and XSB show varying performance with some queries over the same ontology. While XSB is the fastest and consistent system on some queries over LUBM(1), it is among the slowest on other queries. This becomes more pronounced for LUBM(9), where XSB is still quick for a few queries but really slow on others. DLVHex is quicker on LUBM(1) than HEXLite, which shows the contradictory performance on both ontologies.

\subsubsection{LUBM-ext with meta-queries}
It can be seen in Figure 2 that the overall performance of meta-query evaluation is similar to the one in Figure 1. LogicBlox has again the most regular performance in LUBM(1)-ext but shows performance deterioration for LUBM(9)-ext. Clingo and DLV2 have regular performance over varying queries over the same ontology, but are slightly affected by the ontology size. Clingo and DLV2 have shown best performance on a query sq1, next to XSB. Similar comments apply to RDFox and XSB (especially on LUBM(9)-ext), but they are overall slower and performance deteriorates more. XSB shows the best and consistent performance on LUBM(1). NoHR shows somewhat more regular performance here to the one in Figure 1, but it deteriorates with the ontology size and suffers because of the time restriction.

\begin{figure}[!h]
  \centering
  \subfloat[LUBM(1)-ext]{\includegraphics[width=6.2cm]{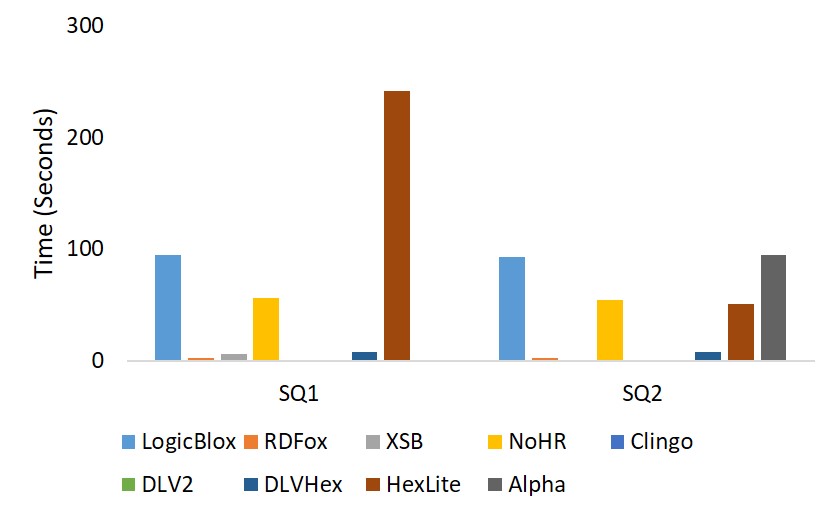}\label{fig:f3}}
  \hfill
  \subfloat[LUBM(9)-ext]{\includegraphics[width=6.2cm]{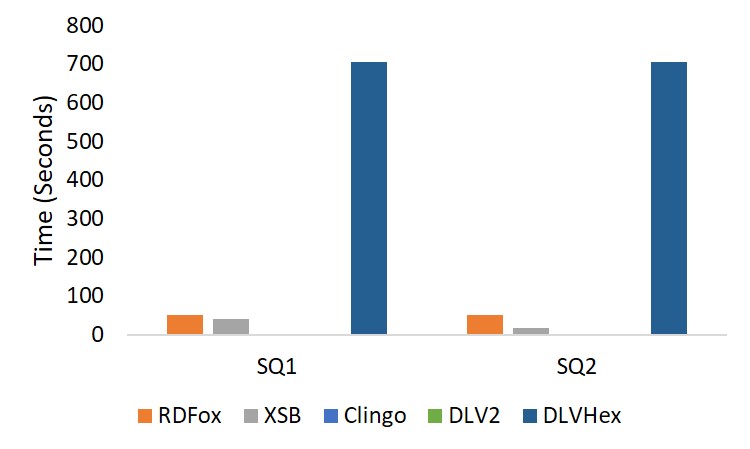}\label{fig:f4}}
  \caption{Experiment 1 LUBM-ext with Meta Queries}
  \label{E1Lext}
\end{figure}

DLVHex shows the same performance on LUBM(1)-ext similar to the results in Figure 1, but its performance gets more affected in LUBM(9)-ext. In contrast, HEXLite, performance suffers from time limits. Alpha  shows more or less the same performance on LUBM(1)-ext than in Figure 1 but was not able to stay within the memory limit on LUBM(9)-ext.
\subsubsection{MODEUS with meta-queries}
The results are reported in Figure 3. Except for mq4 and mq8, for the rest of the queries mq0-mq6, we observe the number of individuals in the ontologies, and intermediate results (determine the next query atom to be executed in each iteration) are quite large. The cause of iterating over all the possible mappings constituted by an \emph{InstanceOf} chain $> 1$ causes an overhead of memory and running time. We can see that most of the tools were not able to compute answers with the given resources. Some tools ran out of memory while loading the data, like LogicBlox and RDFox. In contrast, others consume a lot of memory while grounding atoms like Clingo, DLVHex, HEXLite, and Alpha. NoHR exhibits varied performance, where it successfully computed some queries in the given time, while it ran out of time for others, apparently due to tabling atoms. However, NoHR suffers from quite severe overhead with MODEUS queries compared to the LUBM queries. From the results of Figure 3, one can observe that the queries mq4 and mq8 perform better than others because both queries start with the specific entity and are constituted by an \emph{InstanceOf} chain of length 1.

Other than the resource limitations, there are also some considerations concerning the nature of the MODEUS ontologies:    
\begin{itemize}
    \item As mentioned earlier, the MODEUS dataset consists of meta-layers, which appear to cause many tools to do more inferencing;
    \item the meta-queries consist of different layers of classes, instances and properties, which span over several layers of the dataset; and
    \item we conjecture that the presence of many \textit{disjoint} axioms causes particularly many inferences.
\end{itemize}

\begin{figure}[!h]\label{E1M}
  \centering
  \subfloat[MEF-00]{\includegraphics[width=6.2cm]{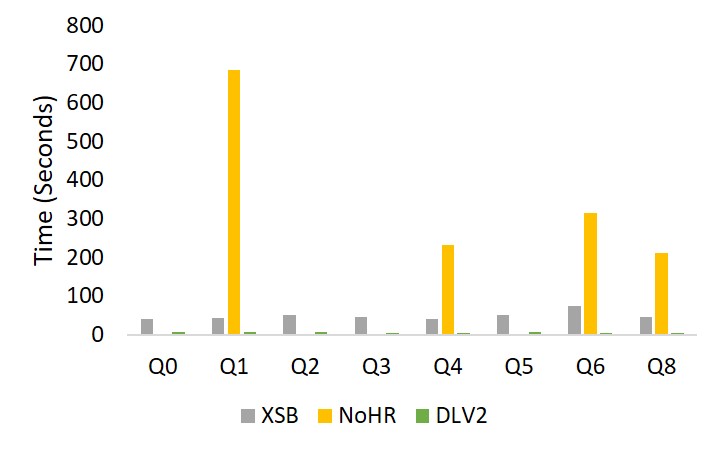}\label{fig:t1}}
  \hfill
  \subfloat[MEF-01]{\includegraphics[width=6.2cm]{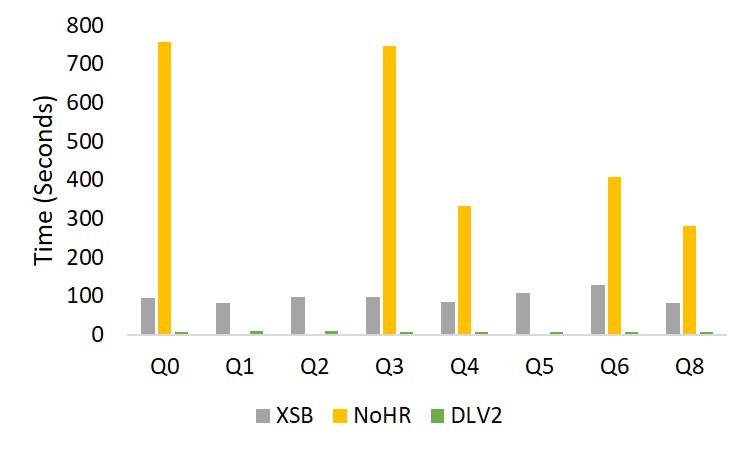}\label{fig:t2}}
  \mbox{}\\
  \subfloat[MEF-02]{\includegraphics[width=6.2cm]{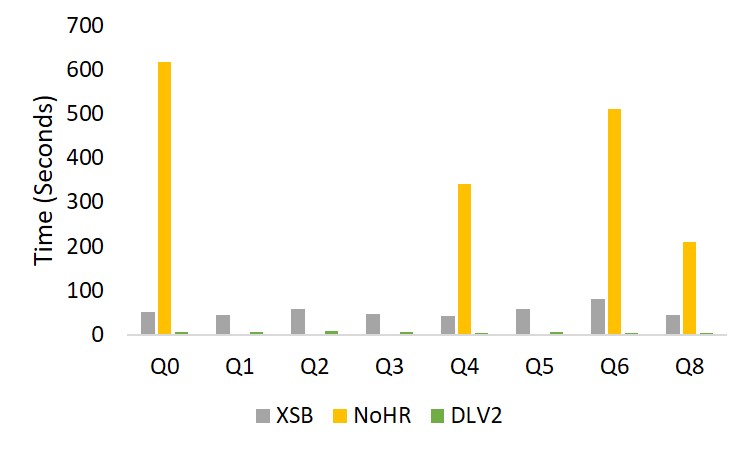}\label{fig:t3}}
  \hfill
  \subfloat[MEF-03]{\includegraphics[width=6.2cm]{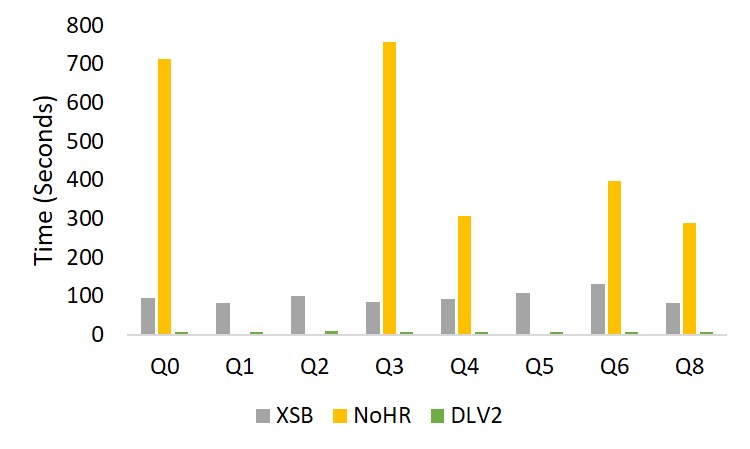}\label{fig:t4}}
  \caption{Experiment 1 MODEUS with Meta Queries}
\end{figure}

XSB and DLV2 show regular performance throughout each dataset. DLV2 is the best overall performer in Figure 3. Its performance is very regular and consistently quick. We assume that the magic set technique implemented in DLV2 has a huge impact here, but it is not clear why LogicBlox (which is also supposed to implement one) cannot profit from it. A possibility is that different sideways information passing strategies (SIPS) are used, which might explain this behavior.

XSB also shows comparatively regular performance but less so than DLV2. It is interesting to see that XSB cannot profit more from its top-down strategy; it is possible that tabling is actually slowing it down in these examples.

\subsection{Experiment 2 -- More Generous Resource Limits}
We next report the results on different datasets under the second experiment settings, allowing for 30 minutes and 128GB  first  LUBM  with standard queries following LUBM-ext with meta-queries, then moving towards MODEUS with meta-queries.

\subsubsection{LUBM with standard queries and simple meta-queries}
\begin{figure}[!h]
  \centering
  \subfloat[LUBM(1)]{\includegraphics[width=6.2cm]{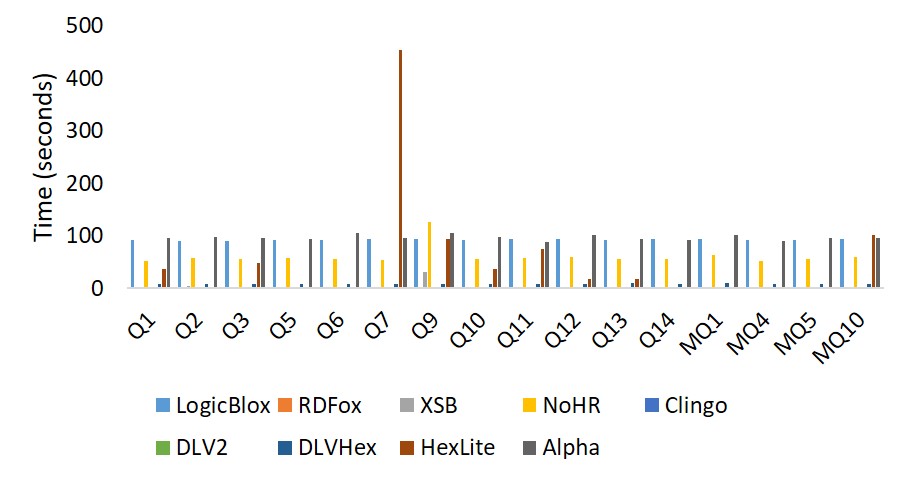}\label{fig:f11}}
  \hfill
  \subfloat[LUBM(9)]{\includegraphics[width=6.2cm]{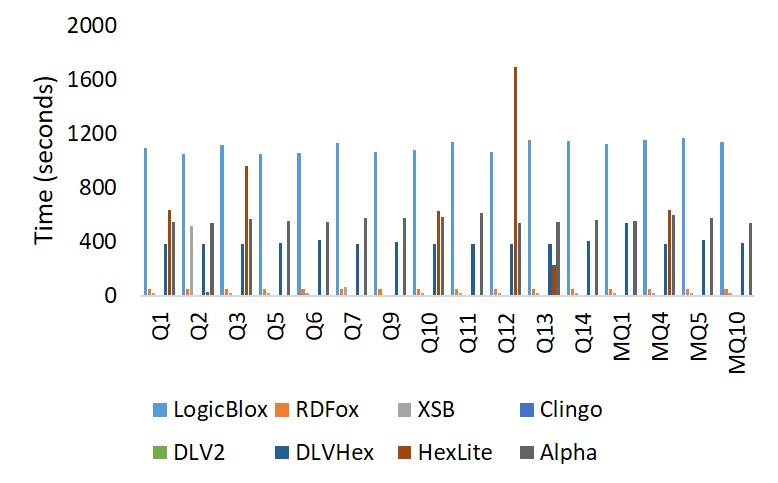}\label{fig:f22}}
  \caption{Experiment 2 LUBM with Standard Queries and Simple Meta-Queries}
\end{figure}

It can be seen in Figure~\ref{fig:f11} and Figure~\ref{fig:f22}, that almost all tools were able to compute answers with the given increased resources, expect for XSB with one outlier (q9) that ran out-of-time during tabling, HexLite with five outliers (q5, q6, q14, mq1, and mq5) in LUBM(1), nine (q5, q6, q7, q9, q11, q14, mq1, mq5, and mq10) in LUBM(9), all of which were out-Of-Time during the search phase, and NoHR ran out-of-time during pre-processing of all queries in LUBM(9). 

DLVHex exhibits the same performance as in Experiment 1 in Figure~\ref{E1L}, whereas some queries in HexLite still exceeded the time limits. Alpha has shown the same performance as in Experiment 1 in Figure~\ref{E1L} on LUBM(1) and is able to meet the memory limit on LUBM(9), but the performance was not particularly satisfactory.

Experiment 2 confirmed the observations for most tools of experiment 1. In particular, XSB, DLV2, and Clingo show consistent performance, where XSB outperforms every single tool in LUBM(1) but is affected by the ontology size in LUBM(9); on the other hand, Clingo and DLV2 show regular performance across both ontologies with a slight increase in time for LUBM(9). 

LogicBlox benefits significantly from increased resources. It now shows consistent performance for both LUBM(1) and LUBM(9). However, it can be seen that its performance deteriorates with the size of the ontology. RDFox shows good performance in LUBM(1) but not so much for LUBM(9). Nevertheless, the overall performance of RDFox across both LUBM(1) and LUBM(9) in both experiments is consistent, which is quite remarkable.

\subsubsection{LUBM-ext with meta-queries}
\begin{figure}[!h]
  \centering
  \subfloat[LUBM(1)-ext]{\includegraphics[width=6.2cm]{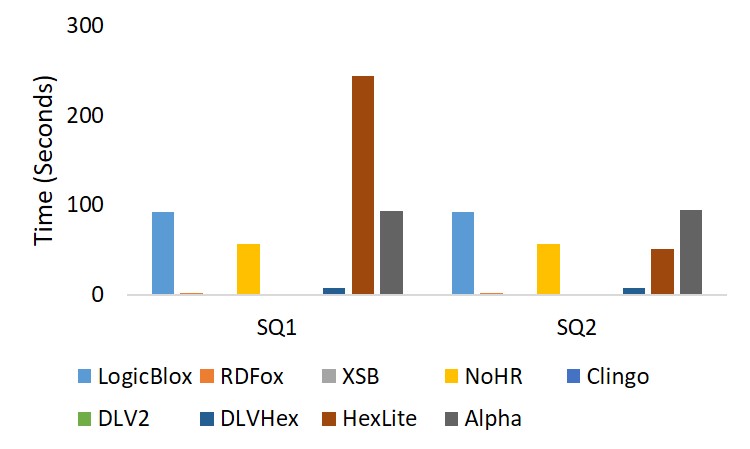}\label{fig:f33}}
  \hfill
  \subfloat[LUBM(9)-ext]{\includegraphics[width=6.2cm]{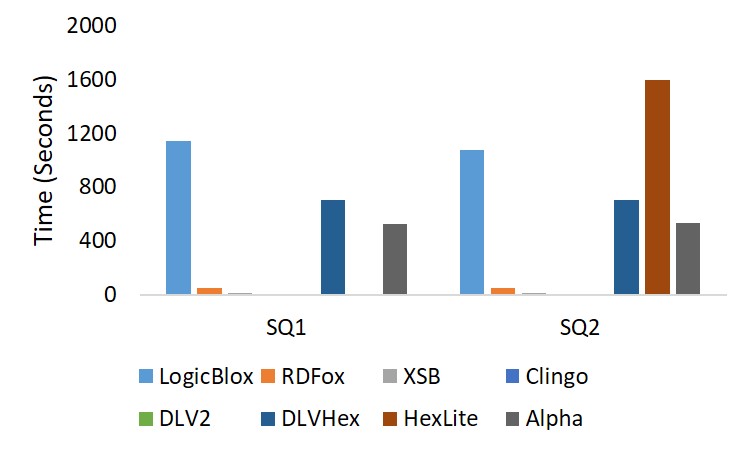}\label{fig:f44}}
  \caption{Experiment 2 LUBM-ext with Meta Queries}
\end{figure}

It can be seen in  Figure~\ref{fig:f33} and Figure~\ref{fig:f44}, that HexLite ran out-of-time during the execution of sq1 and NoHR ran out-of-time during the execution of sq1 and sq2 in LUBM(9)-ext. Both tools, HexLite and NoHR performance greatly affected by the size of the ontologies. Other than the previous two, Alpha performance get very much affected by the ontology size. However, it is able to evaluate both queries but not the quickest.

LogicBlox, performance deteriorates with the extension and we have observed that the "\textit{disjoint-axiom}" has a significant impact on the performance as it seems to cause LogicBlox to check and apply many alternatives. RDFox shows good performance in LUBM(1)-ext but its performance is affected by the ontology size in LUBM(9)-ext. But overall the RDFox performance is satisfactory on the LUBM-ext. 

XSB, Clingo and DLV2 are quickest on LUBM(1)-ext and exhibit reqular performance. However, XSB performance is slightly affected by the size of the input ontology but still performs better than RDFox. Clingo and DLV2 show the most regular and consistent performance over both ontologies and DLV2 outperforms all the tools.

\subsubsection{MODEUS with meta-queries}
In Experiment 2, we get some interesting results compared to Experiment 1. Firstly, XSB, DLV2, and HexLite show a similar performance. Secondly, the tools that broke the memory limits in Experiment 1 like LogicBlox, DLVHex, and Alpha also suffered from the time limit in this experiment. Thirdly, NoHR that could only answer some queries in Experiment 1 was able to answer almost all queries over all four MODUES ontologies except for q2 in MEF-00, MEF-02, MEF-03 and q2,q5 in MEF-01.  

\begin{figure}[!h]
  \centering
  \subfloat[MEF-00]{\includegraphics[width=6.2cm]{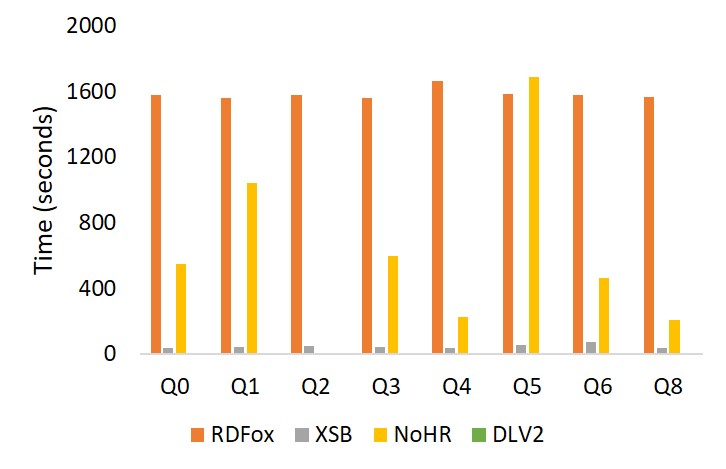}\label{fig:t11}}
  \hfill
  \subfloat[MEF-01]{\includegraphics[width=6.2cm]{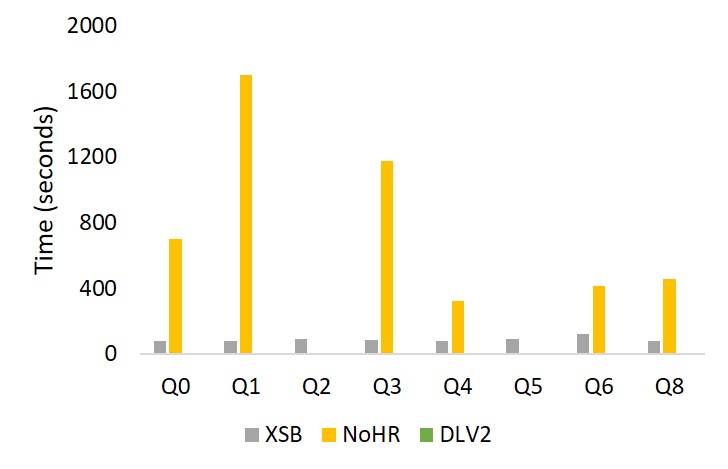}\label{fig:t22}}
  \mbox{}\\
  \subfloat[MEF-02]{\includegraphics[width=6.2cm]{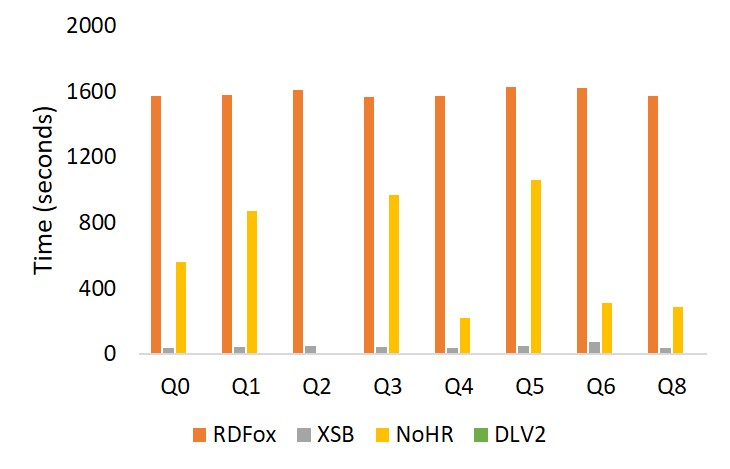}\label{fig:t33}}
  \hfill
  \subfloat[MEF-03]{\includegraphics[width=6.2cm]{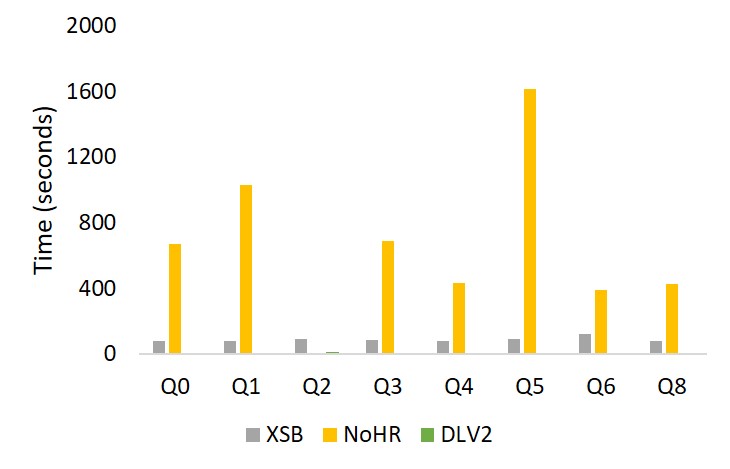}\label{fig:t44}}
  \caption{Experiment 2 MODEUS with Meta Queries}
\end{figure}

Fourthly, RDFox shows its performance on MEF-00 and MEF-02 and suffers from time limitations on the remaining two ontologies. However, the performance on the former two ontologies was not the best performance. Lastly, Clingo produces an error while grounding, indicating that the system ran out of identifiers for ground atoms. Apparently, the problem gives rise to a ground program that is too large for Clingo to handle.

DLV2 and XSB show the most consistent performance across all experiments. However, DLV2 outperforms every single tool in both experiments. The magic set technique appears to play a significant role in the performance of DLV2. On the other hand, the tabling and top-down querying approach have played their role in the consistent performance of XSB across different ontologies and experiments.

\section{Discussion and Future Work}
In this paper we provided both a brief background and an evaluation of a method for answering SPARQL queries over OWL 2 QL ontologies under the metamodeling  semantics  via  a  reduction  to  Datalog  queries. The main contribution of our paper is a performance analysis of various tools that support Datalog, significantly expanding the preliminary analysis in \cite{cima2017sparql}: on the one hand, we additionally consider the MODEUS ontologies that involve meta-axioms, and on the other hand we evaluate significantly more tools of different backgrounds that support Datalog query answering. 

Indeed, our experiments show that especially DLV2, but also XSB appear to be promising back-ends for meta-querying over OWL 2 QL ontologies containing meta-modeling. Surprisingly, neither RDFox nor LogicBlox coped very well with the MODEUS ontologies that involve meta-modeling, while LogicBlox performed very well on LUBM(1) queries. Also, RDFox does not seem to be a good fit for this particular kind of application. Also, the overhead created by NoHR appears quite prohibitive in this scenario. Lastly, Clingo, Alpha, DLVHex, and HexLite do not seem to be well-suited for query answering of this particular kind.

Next, we provide our observations on some tools and try to explain why there is a difference between their performance.

Contrasted with the best in class grounding and solving frameworks like Clingo, Alpha is not yet on par when the grounding is relatively simple to construct, and search is comparatively complex. In our benchmark, we would have expected Alpha to perform well, as there is little search to be done. However, it seems that Alpha is not really geared towards query answering scenarios. We believe that there is potential for Alpha to become a dedicated quey-answering engine as well.

On the other hand, the limits of Clingo's architecture became apparent with the MODEUS ontologies. Clingo integrates the grounder gringo and the solver clasp into a monolithic system to offer more control over the grounding and solving process. Clasp has an inbuilt limit of $2^{28}$ atoms, triggering an error with the MODEUS ontologies, while gringo can handle up to $2^{30}$ atoms. It is apparently not possible to simply increase these limits. Clingo could benefit from a magic set technique as employed by DLV2, which would avoid these size issues.

Compared to the state of the art HEX solver system DLVHex, HexLite is not yet competitive on problems of external atoms computations. HexLite appears to pass over all the optimization tasks (to check the consistency of assignments) to the backend solver Clingo and, as discussed above, Clingo has an inbuilt limit for ground atoms.

DLVHex, which implements the complete Hex language shows quite poor performance on the MODEUS ontologies; it performs a lot of analysis (creating guesses for all external atom even if they are not relevant), bookkeeping and additional computation for optimization to deal with all occurrences of combinations of external computations, which makes DLVHex unnecessarily slow in MODEUS ontologies.

Related to this work are \cite{poggi2016sparql,kontchakov2014answering,gottlob2015beyond}, which consider OWL 2 QL ontologies under DSER and show that it is decidable and can be reduced to Datalog evaluation. However, those works do not consider meta-modeling and meta-querying.

In \cite{lenzerini2016answering,lenzerini2015higher}, the authors overcome the limitations of querying over OWL 2 QL ontologies by introducing the Meta-modeling Semantics MS and show that their proposed algorithm tackles the untyped querying problem in PTime w.r.t.\ data and ontology complexity. The proposed algorithm is implemented in MQ-Mastro\footnote{\url{http://www.modeus.uniroma1.it/modeus/node/6}}, an OWL 2 QL reasoner. It allows for meta-querying over meta-modeling ontologies by offering two types of query evaluation algorithms, NAIVE and Lazy Meta-Grounding (LMG) \cite{lenzerini2020metaquerying}. The system comes with a Graphical User Interface (GUI). Unfortunately, we did not succeed to run it in batch mode, but loading and pre-processing took several minutes, while
answering queries was quick. Overall, however, it cannot match performance of the best Datalog systems. 

As future work we intend to further examine hybrid reasoning for answering meta-queries. The main idea would be to use the same translation function $\tau$ and inference rules, but to keep the non-meta-modeling parts of the ontology  untranslated. This would allow for the use of standard ontology reasoners for the latter part and could lead to performance benefits. The work presented in this paper would be an extreme case of this setting, in which the untranslated part is always empty.
\bibliography{references}

\newpage
\pagestyle{empty}

\appendix
\section{Raw Tables}

\begin{table}[!ht]
\centering
\caption{Experiment1-LUBM with standard-queries (execution times in seconds)}
\label{ex1lubmStable}
\scalebox{1.0}{
\begin{tabular}{*{13}{|@{}c@{}}|}
\noalign{\hrule height 0.5pt}
 & \textbf{q1} & \textbf{q2} & \textbf{q3} & \textbf{q5} & \textbf{q6} & \textbf{q7} & \textbf{q9} & \textbf{q10} & \textbf{q11} & \textbf{q12} & \textbf{q13} & \textbf{q14} \\ 
\noalign{\hrule height 0.5pt}
\multicolumn{13}{|c|}{\textbf{LUBM(1)}} \\ 
\noalign{\hrule height 0.5pt}
\textbf{LogicBlox} & 91.63 & 92.35 & 93.63 & 93.26 & 91.37 & 93.38 & 90.98 & 91.92 & 94.82 & 91.84 & 92.10 & 92.65 \\ 
\noalign{\hrule height 0.5pt}
\textbf{RDFox} & 2.350 & 3.370 & 2.380 & 2.370 & 2.360 & 2.380 & 2.390 & 2.360 & 2.580 & 2.360 & 2.380 & 2.380 \\ 
\noalign{\hrule height 0.5pt}
\textbf{XSB} & \textbf{0.070} & 4.160 & \textbf{0.070} & \textbf{0.070} & \textbf{0.070} & 0.800 & 32.500 & \textbf{0.070} & \textbf{0.070} & \textbf{0.070} & \textbf{0.070} & \textbf{0.070} \\ 
\noalign{\hrule height 0.5pt}
\textbf{NoHR} & 52.82 & 66.710 & 52.650 & 59.650 & 59.670 & 60.430 & 88.010 & 55.500 & 53.700 & 57.300 & 50.690 & 50.600 \\ 
\noalign{\hrule height 0.5pt}
\textbf{Clingo} & 0.110 & \textbf{0.110} & 0.110 & 0.110 & 0.110 & \textbf{0.110} & \textbf{0.110} & 0.100 & 0.110 & 0.110 & 0.100 & 0.110 \\ 
\noalign{\hrule height 0.5pt}
\textbf{DLV2} & 0.100 & 0.120 & 0.110 & 0.110 & 0.110 & \textbf{0.110} & \textbf{0.110} & 0.110 & 0.110 & 0.110 & 0.110 & 0.110 \\
\noalign{\hrule height 0.5pt}
\textbf{DLVHex} & 7.250 & 7.240 & 7.250 & 7.780 & 7.810 & 7.230 & 7.780 & 7.290 & 7.240 & 7.200 & 7.250 & 7.840 \\ 
\noalign{\hrule height 0.5pt}
\textbf{HexLite} & 36.290 & 2.950 & 49.720 & OOT & OOT & 447.290 & 94.100 & 36.15 & 92.330 & 16.860 & 16.880 & OOT \\ 
\noalign{\hrule height 0.5pt}
\textbf{Alpha} & 94.280 & 91.010 & 97.280 & 96.020 & 93.310 & 94.670 & 95.760 & 92.940 & 97.740 & 98.220 & 92.910 & 100.040 \\ 
\noalign{\hrule height 0.5pt}
\multicolumn{13}{|c|}{\textbf{LUBM(9)}} \\ 
\noalign{\hrule height 0.5pt}
\textbf{LogicBlox} & OOM & OOM & OOM & OOM & OOM & OOM & OOM & OOM & OOM & OOM & OOM & OOM \\ 
\noalign{\hrule height 0.5pt}
\textbf{RDFox} & 51.580 & 50.600 & 49.530 & 50.610 & 51.610 & 51.620 & 50.540 & 49.520 & 50.590 & 49.460 & 50.460 & 50.470 \\ 
\noalign{\hrule height 0.5pt}
\textbf{XSB} & 45.770 & 526.140 & 18.270 & 18.160 & 18.230 & 69.830 & OOT & 18.650 & 18.350 & 18.300 & 18.330 & 17.750 \\ 
\noalign{\hrule height 0.5pt}
\textbf{NoHR} & OOT & OOT & OOT & OOT & OOT & OOT & OOT & OOT & OOT & OOT & OOT & OOT \\ 
\noalign{\hrule height 0.5pt}
\textbf{Clingo} & 1.320 & 1.320 & 1.320 & 1.330 & 1.320 & 1.320 & 1.330 & 1.330 & 1.330 & 1.330 & 1.330 & 1.320 \\ 
\noalign{\hrule height 0.5pt}
\textbf{DLV2} & \textbf{0.980} & \textbf{1.090} & \textbf{0.860} & \textbf{1.120} & \textbf{1.070} & \textbf{1.090} & \textbf{1.090} & \textbf{1.100} & \textbf{1.090} & \textbf{1.080} & \textbf{1.090} & \textbf{1.050} \\ 
\noalign{\hrule height 0.5pt}
\textbf{DLVHex} & 386.320 & 387.790 & 385.880 & 389.610 & 412.470 & 386.750 & 386.230 & 384.920 & 387.130 & 386.650 & 386.860 & 386.860 \\ 
\noalign{\hrule height 0.5pt}
\textbf{HexLite} & 646.540 & 30.020 & OOT & OOT & OOT & OOT & OOT & 636.570 & OOT & OOT & 221.820 & OOT \\ 
\noalign{\hrule height 0.5pt}
\textbf{Alpha} & OOM & OOM & OOM & OOM & OOM & OOM & OOM & OOM & OOM & OOM & OOM & OOM \\
\noalign{\hrule height 0.5pt}
\end{tabular}}
\end{table}

\begin{table}
\centering
\caption{Experiment1-LUBM with simple meta-queries (execution times in seconds)}
\label{ex1lubmMtable}
\begin{tabular}{|c|c|c|c|c|} 
\noalign{\hrule height 0.5pt}
                   & \textbf{mq1}   & \textbf{mq4}   & \textbf{mq5}   & \textbf{mq10}   \\ 
\noalign{\hrule height 0.5pt}
\multicolumn{5}{|c|}{\textbf{LUBM(1)}}                                                  \\ 
\noalign{\hrule height 0.5pt}
\textbf{LogicBlox} & 92.99          & 93.07          & 92.88          & 92.13           \\ 
\noalign{\hrule height 0.5pt}
\textbf{RDFox}     & 2.840          & 2.370          & 2.370          & 2.360           \\ 
\noalign{\hrule height 0.5pt}
\textbf{XSB}       & 0.190          & \textbf{0.000} & \textbf{0.070} & \textbf{0.070}  \\ 
\noalign{\hrule height 0.5pt}
\textbf{NoHR}      & 65.300         & 49.530         & 54.690         & 54.000          \\ 
\noalign{\hrule height 0.5pt}
\textbf{Clingo}    & \textbf{0.110} & 0.110          & 0.110          & 0.110           \\ 
\noalign{\hrule height 0.5pt}
\textbf{DLV2}      & \textbf{0.110} & 0.110          & 0.110          & 0.110           \\ 
\noalign{\hrule height 0.5pt}
\textbf{DLVHex}    & 9.890          & 7.210          & 7.770          & 7.230           \\ 
\noalign{\hrule height 0.5pt}
\textbf{HexLite}   & OOT            & 2.980          & OOT            & 100.580         \\ 
\noalign{\hrule height 0.5pt}
\textbf{Alpha}     & 93.460         & 95.260         & 96.510         & 100.760         \\ 
\noalign{\hrule height 0.5pt}
\multicolumn{5}{|c|}{\textbf{LUBM(9)}}                                                  \\ 
\noalign{\hrule height 0.5pt}
\textbf{LogicBlox} & OOM            & OOM            & OOM            & OOM             \\ 
\noalign{\hrule height 0.5pt}
\textbf{RDFox}     & 49.410         & 50.640         & 52.700         & 50.570          \\ 
\noalign{\hrule height 0.5pt}
\textbf{XSB}       & 18.770         & 18.300         & 19.750         & 18.800          \\ 
\noalign{\hrule height 0.5pt}
\textbf{NoHR}      & OOT            & OOT            & OOT            & OOT             \\ 
\noalign{\hrule height 0.5pt}
\textbf{Clingo}    & 1.330          & 1.320          & 1.330          & 1.330           \\ 
\noalign{\hrule height 0.5pt}
\textbf{DLV2}      & \textbf{1.150} & \textbf{1.090} & \textbf{1.130} & \textbf{1.090}  \\ 
\noalign{\hrule height 0.5pt}
\textbf{DLVHex}    & 534.930        & 387.270        & 407.170        & 386.420         \\ 
\noalign{\hrule height 0.5pt}
\textbf{HexLite}   & OOT            & 632.930        & OOT            & OOT             \\ 
\noalign{\hrule height 0.5pt}
\textbf{Alpha}     & OOM            & OOM            & OOM            & OOM             \\
\noalign{\hrule height 0.5pt}
\end{tabular}
\end{table}

\begin{table}
\centering
\caption{Experiment1-LUBM-ext with meta-queries (execution times in seconds)}
\label{ex1lubmMtable}
\begin{tabular}{|c|c|c|} 
\noalign{\hrule height 0.5pt}
                   & \textbf{sq1}   & \textbf{sq2}    \\ 
\noalign{\hrule height 0.5pt}
\multicolumn{3}{|c|}{\textbf{LUBM(1)-ext}}                \\ 
\noalign{\hrule height 0.5pt}
\textbf{LogicBlox} & 94.82          & 93.00           \\ 
\noalign{\hrule height 0.5pt}
\textbf{RDFox}     & 2.370          & 2.390           \\ 
\noalign{\hrule height 0.5pt}
\textbf{XSB}       & 6.000          & \textbf{0.080}  \\ 
\noalign{\hrule height 0.5pt}
\textbf{NoHR}      & 56.290         & 54.590          \\ 
\noalign{\hrule height 0.5pt}
\textbf{Clingo}    & \textbf{0.110} & 0.110           \\ 
\noalign{\hrule height 0.5pt}
\textbf{DLV2}      & 0.120          & 0.110           \\ 
\noalign{\hrule height 0.5pt}
\textbf{DLVHex}    & 7.750          & 7.760           \\ 
\noalign{\hrule height 0.5pt}
\textbf{HexLite}   & 241.700        & 50.720          \\ 
\noalign{\hrule height 0.5pt}
\textbf{Alpha}     & 96.290         & 94.200          \\ 
\noalign{\hrule height 0.5pt}
\multicolumn{3}{|c|}{\textbf{LUBM(9)-ext}}                \\ 
\noalign{\hrule height 0.5pt}
\textbf{LogicBlox} & OOM            & OOM             \\ 
\noalign{\hrule height 0.5pt}
\textbf{RDFox}     & 50.530         & 49.540          \\ 
\noalign{\hrule height 0.5pt}
\textbf{XSB}       & 40.680         & 18.130          \\ 
\noalign{\hrule height 0.5pt}
\textbf{NoHR}      & OOT            & OOT             \\ 
\noalign{\hrule height 0.5pt}
\textbf{Clingo}    & 1.320          & 1.320           \\ 
\noalign{\hrule height 0.5pt}
\textbf{DLV2}      & \textbf{1.100} & \textbf{0.980}  \\ 
\noalign{\hrule height 0.5pt}
\textbf{DLVHex}    & 704.490        & 704.750         \\ 
\noalign{\hrule height 0.5pt}
\textbf{HexLite}   & OOT            & OOT             \\ 
\noalign{\hrule height 0.5pt}
\textbf{Alpha}     & OOM            & OOM             \\
\noalign{\hrule height 0.5pt}
\end{tabular}
\end{table}

\begin{table}[h!]
\centering
\caption{Experiment1-MODEUS with meta-queries (execution times in seconds)}
\label{ex1modustable}
\scalebox{1.0}{
\begin{tabular}{*{9}{|@{}c@{}}|}
\noalign{\hrule height 0.5pt}
\multicolumn{1}{|l|}{}                   & \multicolumn{1}{l|}{\textbf{mq0}} & \multicolumn{1}{l|}{\textbf{mq1}} & \multicolumn{1}{l|}{\textbf{mq2}} & \multicolumn{1}{l|}{\textbf{mq3}} & \multicolumn{1}{l|}{\textbf{mq4}} & \multicolumn{1}{l|}{\textbf{mq5}} & \multicolumn{1}{l|}{\textbf{mq6}} & \textbf{mq8} \\ \noalign{\hrule height 0.5pt}
\multicolumn{9}{|c|}{\textbf{MEF-00}}                                                                                 \\ \noalign{\hrule height 0.5pt}
\multicolumn{1}{|l|}{\textbf{LogicBlox}} & \multicolumn{1}{l|}{OOM}          & \multicolumn{1}{l|}{OOM}          & \multicolumn{1}{l|}{OOM}          & \multicolumn{1}{l|}{OOM}          & \multicolumn{1}{l|}{OOM}          & \multicolumn{1}{l|}{OOM}          & \multicolumn{1}{l|}{OOM}          & OOM          \\ \noalign{\hrule height 0.5pt}
\multicolumn{1}{|l|}{\textbf{RDFox}}     & \multicolumn{1}{l|}{OOM}          & \multicolumn{1}{l|}{OOM}          & \multicolumn{1}{l|}{OOM}          & \multicolumn{1}{l|}{OOM}          & \multicolumn{1}{l|}{OOM}          & \multicolumn{1}{l|}{OOM}          & \multicolumn{1}{l|}{OOM}          & OOM          \\ \noalign{\hrule height 0.5pt}
\multicolumn{1}{|l|}{\textbf{XSB}}       & \multicolumn{1}{l|}{39.750}       & \multicolumn{1}{l|}{41.910}       & \multicolumn{1}{l|}{51.300s}      & \multicolumn{1}{l|}{45.130}       & \multicolumn{1}{l|}{40.020}       & \multicolumn{1}{l|}{51.390}       & \multicolumn{1}{l|}{74.040}       & 44.440       \\ \noalign{\hrule height 0.5pt}
\multicolumn{1}{|l|}{\textbf{NoHR}}      & \multicolumn{1}{l|}{OOT}          & \multicolumn{1}{l|}{682.770}      & \multicolumn{1}{l|}{OOM}          & \multicolumn{1}{l|}{OOT}          & \multicolumn{1}{l|}{232.580}      & \multicolumn{1}{l|}{OOT}          & \multicolumn{1}{l|}{314.250}      & 210.97       \\ \noalign{\hrule height 0.5pt}
\multicolumn{1}{|l|}{\textbf{Clingo}}    & \multicolumn{1}{l|}{OOM}          & \multicolumn{1}{l|}{OOM}          & \multicolumn{1}{l|}{OOM}          & \multicolumn{1}{l|}{OOM}          & \multicolumn{1}{l|}{OOM}          & \multicolumn{1}{l|}{OOM}          & \multicolumn{1}{l|}{OOM}          & OOM          \\ \noalign{\hrule height 0.5pt}
\multicolumn{1}{|l|}{\textbf{DLV2}}       & \multicolumn{1}{l|}{\textbf{5.450}}        & \multicolumn{1}{l|}{\textbf{5.870}}        & \multicolumn{1}{l|}{\textbf{7.570}}        & \multicolumn{1}{l|}{\textbf{5.290}}        & \multicolumn{1}{l|}{\textbf{4.750}}        & \multicolumn{1}{l|}{\textbf{6.700}}        & \multicolumn{1}{l|}{\textbf{5.040}}    &\multicolumn{1}{l|}{\textbf{ 4.780}}        \\ \noalign{\hrule height 0.5pt}
\multicolumn{1}{|l|}{\textbf{DLVHex}}    & \multicolumn{1}{l|}{OOM}          & \multicolumn{1}{l|}{OOM}          & \multicolumn{1}{l|}{OOM}          & \multicolumn{1}{l|}{OOM}          & \multicolumn{1}{l|}{OOM}          & \multicolumn{1}{l|}{OOM}          & \multicolumn{1}{l|}{OOM}          & OOM          \\ \noalign{\hrule height 0.5pt}
\multicolumn{1}{|l|}{\textbf{HexLite}}   & \multicolumn{1}{l|}{OOM}          & \multicolumn{1}{l|}{OOM}          & \multicolumn{1}{l|}{OOM}          & \multicolumn{1}{l|}{OOM}          & \multicolumn{1}{l|}{OOM}          & \multicolumn{1}{l|}{OOM}          & \multicolumn{1}{l|}{OOM}          & OOM          \\ \noalign{\hrule height 0.5pt}
\multicolumn{1}{|l|}{\textbf{Alpha}}     & \multicolumn{1}{l|}{OOM}          & \multicolumn{1}{l|}{OOM}          & \multicolumn{1}{l|}{OOM}          & \multicolumn{1}{l|}{OOM}          & \multicolumn{1}{l|}{OOM}          & \multicolumn{1}{l|}{OOM}          & \multicolumn{1}{l|}{OOM}          & OOM          \\ \noalign{\hrule height 0.5pt}
\multicolumn{9}{|c|}{\textbf{MEF-01}}                                                                 \\ \noalign{\hrule height 0.5pt}
\multicolumn{1}{|l|}{\textbf{LogicBlox}} & \multicolumn{1}{l|}{OOM}          & \multicolumn{1}{l|}{OOM}          & \multicolumn{1}{l|}{OOM}          & \multicolumn{1}{l|}{OOM}          & \multicolumn{1}{l|}{OOM}          & \multicolumn{1}{l|}{OOM}          & \multicolumn{1}{l|}{OOM}          & OOM          \\ \noalign{\hrule height 0.5pt}
\multicolumn{1}{|l|}{\textbf{RDFox}}     & \multicolumn{1}{l|}{OOM}          & \multicolumn{1}{l|}{OOM}          & \multicolumn{1}{l|}{OOM}          & \multicolumn{1}{l|}{OOM}          & \multicolumn{1}{l|}{OOM}          & \multicolumn{1}{l|}{OOM}          & \multicolumn{1}{l|}{OOM}          & OOM          \\ \noalign{\hrule height 0.5pt}
\multicolumn{1}{|l|}{\textbf{XSB}}       & \multicolumn{1}{l|}{94.100}       & \multicolumn{1}{l|}{82.860}       & \multicolumn{1}{l|}{97.160}       & \multicolumn{1}{l|}{95.890}       & \multicolumn{1}{l|}{84.470}       & \multicolumn{1}{l|}{108.490}      & \multicolumn{1}{l|}{128.750}      & 80.440       \\ \noalign{\hrule height 0.5pt}
\multicolumn{1}{|l|}{\textbf{NoHR}}      & \multicolumn{1}{l|}{757.19}       & \multicolumn{1}{l|}{OOT}          & \multicolumn{1}{l|}{OOM}          & \multicolumn{1}{l|}{745.890}      & \multicolumn{1}{l|}{331.340}      & \multicolumn{1}{l|}{OOT}          & \multicolumn{1}{l|}{406.360}      & 280.880      \\ \noalign{\hrule height 0.5pt}
\multicolumn{1}{|l|}{\textbf{Clingo}}    & \multicolumn{1}{l|}{OOM}          & \multicolumn{1}{l|}{OOM}          & \multicolumn{1}{l|}{OOM}          & \multicolumn{1}{l|}{OOM}          & \multicolumn{1}{l|}{OOM}          & \multicolumn{1}{l|}{OOM}          & \multicolumn{1}{l|}{OOM}          & OOM          \\ \noalign{\hrule height 0.5pt}
\multicolumn{1}{|l|}{\textbf{DLV2}}       & \multicolumn{1}{l|}{\textbf{6.400}}        & \multicolumn{1}{l|}{\textbf{8.060}}        & \multicolumn{1}{l|}{\textbf{9.740}}        & \multicolumn{1}{l|}{\textbf{6.540}}        & \multicolumn{1}{l|}{\textbf{5.750}}        & \multicolumn{1}{l|}{\textbf{7.520}}        & \multicolumn{1}{l|}{\textbf{5.630}}        & \textbf{5.530}        \\ \noalign{\hrule height 0.5pt}
\multicolumn{1}{|l|}{\textbf{DLVHex}}    & \multicolumn{1}{l|}{OOM}          & \multicolumn{1}{l|}{OOM}          & \multicolumn{1}{l|}{OOM}          & \multicolumn{1}{l|}{OOM}          & \multicolumn{1}{l|}{OOM}          & \multicolumn{1}{l|}{OOM}          & \multicolumn{1}{l|}{OOM}          & OOM          \\ \noalign{\hrule height 0.5pt}
\multicolumn{1}{|l|}{\textbf{HexLite}}   & \multicolumn{1}{l|}{OOM}          & \multicolumn{1}{l|}{OOM}          & \multicolumn{1}{l|}{OOM}          & \multicolumn{1}{l|}{OOM}          & \multicolumn{1}{l|}{OOM}          & \multicolumn{1}{l|}{OOM}          & \multicolumn{1}{l|}{OOM}          & OOM          \\ \noalign{\hrule height 0.5pt}
\multicolumn{1}{|l|}{\textbf{Alpha}}     & \multicolumn{1}{l|}{OOM}          & \multicolumn{1}{l|}{OOM}          & \multicolumn{1}{l|}{OOM}          & \multicolumn{1}{l|}{OOM}          & \multicolumn{1}{l|}{OOM}          & \multicolumn{1}{l|}{OOM}          & \multicolumn{1}{l|}{OOM}          & OOM          \\ \noalign{\hrule height 0.5pt}
\multicolumn{9}{|c|}{\textbf{MEF-02}}                                                                                                    \\ \noalign{\hrule height 0.5pt}
\multicolumn{1}{|l|}{\textbf{LogicBlox}} & \multicolumn{1}{l|}{OOM}          & \multicolumn{1}{l|}{OOM}          & \multicolumn{1}{l|}{OOM}          & \multicolumn{1}{l|}{OOM}          & \multicolumn{1}{l|}{OOM}          & \multicolumn{1}{l|}{OOM}          & \multicolumn{1}{l|}{OOM}          & OOM          \\ \noalign{\hrule height 0.5pt}
\multicolumn{1}{|l|}{\textbf{RDFox}}     & \multicolumn{1}{l|}{OOM}          & \multicolumn{1}{l|}{OOM}          & \multicolumn{1}{l|}{OOM}          & \multicolumn{1}{l|}{OOM}          & \multicolumn{1}{l|}{OOM}          & \multicolumn{1}{l|}{OOM}          & \multicolumn{1}{l|}{OOM}          & OOM          \\ \noalign{\hrule height 0.5pt}
\multicolumn{1}{|l|}{\textbf{XSB}}       & \multicolumn{1}{l|}{50.450}       & \multicolumn{1}{l|}{44.500}       & \multicolumn{1}{l|}{58.770}       & \multicolumn{1}{l|}{46.280}       & \multicolumn{1}{l|}{41.160}       & \multicolumn{1}{l|}{57.620}       & \multicolumn{1}{l|}{80.210}       & 45.190       \\ \noalign{\hrule height 0.5pt}
\multicolumn{1}{|l|}{\textbf{NoHR}}      & \multicolumn{1}{l|}{617.270}      & \multicolumn{1}{l|}{OOT}          & \multicolumn{1}{l|}{OOM}          & \multicolumn{1}{l|}{OOT}          & \multicolumn{1}{l|}{340.620}      & \multicolumn{1}{l|}{OOT}          & \multicolumn{1}{l|}{509.190}      & 209.420      \\ \noalign{\hrule height 0.5pt}
\multicolumn{1}{|l|}{\textbf{Clingo}}    & \multicolumn{1}{l|}{OOM}          & \multicolumn{1}{l|}{OOM}          & \multicolumn{1}{l|}{OOM}          & \multicolumn{1}{l|}{OOM}          & \multicolumn{1}{l|}{OOM}          & \multicolumn{1}{l|}{OOM}          & \multicolumn{1}{l|}{OOM}          & OOM          \\ \noalign{\hrule height 0.5pt}
\multicolumn{1}{|l|}{\textbf{DLV2}}       & \multicolumn{1}{l|}{\textbf{5.110}}        & \multicolumn{1}{l|}{\textbf{5.610}}        & \multicolumn{1}{l|}{\textbf{7.410}}        & \multicolumn{1}{l|}{\textbf{5.390}}        & \multicolumn{1}{l|}{\textbf{4.500}}        & \multicolumn{1}{l|}{\textbf{6.520}}        & \multicolumn{1}{l|}{\textbf{4.640}}        & \textbf{4.590}        \\ \noalign{\hrule height 0.5pt}
\multicolumn{1}{|l|}{\textbf{DLVHex}}    & \multicolumn{1}{l|}{OOM}          & \multicolumn{1}{l|}{OOM}          & \multicolumn{1}{l|}{OOM}          & \multicolumn{1}{l|}{OOM}          & \multicolumn{1}{l|}{OOM}          & \multicolumn{1}{l|}{OOM}          & \multicolumn{1}{l|}{OOM}          & OOM          \\ \noalign{\hrule height 0.5pt}
\multicolumn{1}{|l|}{\textbf{HexLite}}   & \multicolumn{1}{l|}{OOM}          & \multicolumn{1}{l|}{OOM}          & \multicolumn{1}{l|}{OOM}          & \multicolumn{1}{l|}{OOM}          & \multicolumn{1}{l|}{OOM}          & \multicolumn{1}{l|}{OOM}          & \multicolumn{1}{l|}{OOM}          & OOM          \\ \noalign{\hrule height 0.5pt}
\multicolumn{1}{|l|}{\textbf{Alpha}}     & \multicolumn{1}{l|}{OOM}          & \multicolumn{1}{l|}{OOM}          & \multicolumn{1}{l|}{OOM}          & \multicolumn{1}{l|}{OOM}          & \multicolumn{1}{l|}{OOM}          & \multicolumn{1}{l|}{OOM}          & \multicolumn{1}{l|}{OOM}          & OOM          \\ \noalign{\hrule height 0.5pt}
\multicolumn{9}{|c|}{\textbf{MEF-03}}                                                                                           \\ \noalign{\hrule height 0.5pt}
\multicolumn{1}{|l|}{\textbf{LogicBlox}} & \multicolumn{1}{l|}{OOM}          & \multicolumn{1}{l|}{OOM}          & \multicolumn{1}{l|}{OOM}          & \multicolumn{1}{l|}{OOM}          & \multicolumn{1}{l|}{OOM}          & \multicolumn{1}{l|}{OOM}          & \multicolumn{1}{l|}{OOM}          & OOM          \\ \noalign{\hrule height 0.5pt}
\multicolumn{1}{|l|}{\textbf{RDFox}}     & \multicolumn{1}{l|}{OOM}          & \multicolumn{1}{l|}{OOM}          & \multicolumn{1}{l|}{OOM}          & \multicolumn{1}{l|}{OOM}          & \multicolumn{1}{l|}{OOM}          & \multicolumn{1}{l|}{OOM}          & \multicolumn{1}{l|}{OOM}          & OOM          \\ \noalign{\hrule height 0.5pt}
\multicolumn{1}{|l|}{\textbf{XSB}}       & \multicolumn{1}{l|}{94.230}       & \multicolumn{1}{l|}{80.740}       & \multicolumn{1}{l|}{99.580}       & \multicolumn{1}{l|}{83.800}       & \multicolumn{1}{l|}{92.660}       & \multicolumn{1}{l|}{107.240}      & \multicolumn{1}{l|}{131.800}      & 82.500       \\ \noalign{\hrule height 0.5pt}
\multicolumn{1}{|l|}{\textbf{NoHR}}      & \multicolumn{1}{l|}{713.410}      & \multicolumn{1}{l|}{OOT}          & \multicolumn{1}{l|}{OOM}          & \multicolumn{1}{l|}{757.010}      & \multicolumn{1}{l|}{306.180}      & \multicolumn{1}{l|}{OOM}          & \multicolumn{1}{l|}{396.000}      & 289.420      \\ \noalign{\hrule height 0.5pt}
\multicolumn{1}{|l|}{\textbf{Clingo}}    & \multicolumn{1}{l|}{OOM}          & \multicolumn{1}{l|}{OOM}          & \multicolumn{1}{l|}{OOM}          & \multicolumn{1}{l|}{OOM}          & \multicolumn{1}{l|}{OOM}          & \multicolumn{1}{l|}{OOM}          & \multicolumn{1}{l|}{OOM}          & OOM          \\ \noalign{\hrule height 0.5pt}
\multicolumn{1}{|l|}{\textbf{DLV2}}       & \multicolumn{1}{l|}{\textbf{6.830}}        & \multicolumn{1}{l|}{\textbf{7.350}}        & \multicolumn{1}{l|}{\textbf{10.010}}       & \multicolumn{1}{l|}{\textbf{6.880}}        & \multicolumn{1}{l|}{\textbf{5.540}}        & \multicolumn{1}{l|}{\textbf{7.250}}        & \multicolumn{1}{l|}{\textbf{5.450}}        & \textbf{5.540}        \\ \noalign{\hrule height 0.5pt}
\multicolumn{1}{|l|}{\textbf{DLVHex}}    & \multicolumn{1}{l|}{OOM}          & \multicolumn{1}{l|}{OOM}          & \multicolumn{1}{l|}{OOM}          & \multicolumn{1}{l|}{OOM}          & \multicolumn{1}{l|}{OOM}          & \multicolumn{1}{l|}{OOM}          & \multicolumn{1}{l|}{OOM}          & OOM          \\ \noalign{\hrule height 0.5pt}
\multicolumn{1}{|l|}{\textbf{HexLite}}   & \multicolumn{1}{l|}{OOM}          & \multicolumn{1}{l|}{OOM}          & \multicolumn{1}{l|}{OOM}          & \multicolumn{1}{l|}{OOM}          & \multicolumn{1}{l|}{OOM}          & \multicolumn{1}{l|}{OOM}          & \multicolumn{1}{l|}{OOM}          & OOM          \\ \noalign{\hrule height 0.5pt}
\multicolumn{1}{|l|}{\textbf{Alpha}}     & \multicolumn{1}{l|}{OOM}          & \multicolumn{1}{l|}{OOM}          & \multicolumn{1}{l|}{OOM}          & \multicolumn{1}{l|}{OOM}          & \multicolumn{1}{l|}{OOM}          & \multicolumn{1}{l|}{OOM}          & \multicolumn{1}{l|}{OOM}          & OOM          \\ \noalign{\hrule height 0.5pt}
\end{tabular}}
\end{table}

\begin{table}[!ht]
    \centering
    \caption{Experiment2-LUBM with standard-queries (execution times in seconds)}
\label{ex2lubmStable}
\scalebox{1.0}{
    \begin{tabular}{*{13}{|@{}c@{}}|}
    \noalign{\hrule height 0.5pt}
        ~ & \textbf{q1} & \textbf{q2} & \textbf{q3} & \textbf{q5} & \textbf{q6} & \textbf{q7} & \textbf{q9} & \textbf{q10} & \textbf{q11} & \textbf{q12} & \textbf{q13} & \textbf{q14} \\ \noalign{\hrule height 0.5pt}
        \noalign{\hrule height 0.5pt}
\multicolumn{13}{|c|}{\textbf{LUBM(1)}} \\ 
\noalign{\hrule height 0.5pt}
        \textbf{LogicBlox} & 92.92 & 90.55 & 90.61 & 92.46 & 92.91 & 94.68 & 94.37 & 92.52 & 93.86 & 93.42 & 92.3 & 94.21  \\ \noalign{\hrule height 0.5pt}
        \textbf{RDFox} & 2.41 & 2.4 & 2.36 & 2.37 & 2.36 & 2.37 & 2.59 & 2.59 & 2.42 & 2.35 & 2.54 & 2.37  \\ \noalign{\hrule height 0.5pt}
        \textbf{XSB} & 0.08 & 4.2 & 0.07 & 0.06 & 0.07 & 0.81 & 31.43 & 0.08 & 0.07 & 0.08 & 0.07 & 0.07  \\ \noalign{\hrule height 0.5pt}
        \textbf{NoHR} & 51.98 & 57.58 & 55.36 & 57.66 & 56.2 & 53.48 & 126.7 & 54.82 & 57.84 & 58.77 & 56.37 & 56.19  \\ \noalign{\hrule height 0.5pt}
        \textbf{Clingo} & \textbf{0.1} & \textbf{0.11} & \textbf{0.11} & \textbf{0.11} & \textbf{0.11} & \textbf{0.11} & \textbf{0.11} & \textbf{0.11} & \textbf{0.11} & \textbf{0.11} & \textbf{0.11} & \textbf{0.11}  \\ \noalign{\hrule height 0.5pt}
        \textbf{DLV2} & 0.12 & 0.11 & 0.11 & 0.12 & 0.11 & 0.11 & 0.12 & 0.12 & 0.11 & 0.11 & 0.11 & 0.12  \\ \noalign{\hrule height 0.5pt}
        \textbf{DLVHex} & 7.3 & 8.84 & 7.26 & 7.75 & 7.77 & 7.26 & 7.25 & 7.26 & 7.83 & 7.2 & 9.88 & 7.84  \\ \noalign{\hrule height 0.5pt}
        \textbf{HexLite} & 36.3 & 2.93 & 48.63 & OOT & OOT & 453.79 & 94.25 & 36.18 & 74.65 & 16.87 & 16.9 & OOT  \\ \noalign{\hrule height 0.5pt}
        \textbf{Alpha} & 96.07 & 96.81 & 96.46 & 94.15 & 104.5 & 95.53 & 105.6 & 97.41 & 88.63 & 102 & 94.14 & 91.79  \\ 
        \noalign{\hrule height 0.5pt}
\multicolumn{13}{|c|}{\textbf{LUBM(9)}} \\ 
\noalign{\hrule height 0.5pt}
        \textbf{LogicBlox} & 1091.11 & 1050.66 & 1116.33 & 1050.21 & 1056.76 & 1130.51 & 1063.32 & 1077.76 & 1135.56 & 1063.66 & 1154.39 & 1142.95  \\ \noalign{\hrule height 0.5pt}
        \textbf{RDFox} & 51.85 & 51.79 & 50.68 & 49.58 & 49.68 & 49.53 & 50.74 & 51.93 & 51.73 & 49.46 & 49.53 & 50.75  \\ \noalign{\hrule height 0.5pt}
        \textbf{XSB} & 16.6 & 519.47 & 17.13 & 17.67 & 17.16 & 66.59 & OOT & 17.25 & 16.52 & 17.69 & 17.08 & 17.18  \\ \noalign{\hrule height 0.5pt}
        \textbf{NoHR} & OOT & OOT & OOT & OOT & OOT & OOT & OOT & OOT & OOT & OOT & OOT & OOT  \\ \noalign{\hrule height 0.5pt}
        \textbf{Clingo} & 1.32 & 1.32 & 1.32 & 1.33 & 1.33 & 1.32 & 1.33 & 1.32 & 1.32 & 1.33 & 1.33 & 1.33  \\ \noalign{\hrule height 0.5pt}
        \textbf{DLV2} & \textbf{0.96} & \textbf{0.97} & \textbf{0.98} & \textbf{0.86} & \textbf{0.96} & \textbf{0.94} & \textbf{0.96} & \textbf{0.98} & \textbf{0.96} & \textbf{0.97} & \textbf{0.96} & \textbf{0.95}  \\ \noalign{\hrule height 0.5pt}
        \textbf{DLVHex} & 386.31 & 384.85 & 384.86 & 388.87 & 412.19 & 385.95 & 397.48 & 385.77 & 386.09 & 384.82 & 384.86 & 406.65  \\ \noalign{\hrule height 0.5pt}
        \textbf{HexLite} & 631.31 & 28.71 & 961.19 & OOT & OOT & OOT & OOT & 630.04 & OOT & 1692.5 & 224.57 & OOT  \\ \noalign{\hrule height 0.5pt}
        \textbf{Alpha} & 547.32 & 534.45 & 566.16 & 554.59 & 542.26 & 572.3 & 577.5 & 582.69 & 610.61 & 540.2 & 544.85 & 558.06  \\ \noalign{\hrule height 0.5pt}
    \end{tabular}}
\end{table}

\begin{table}
\centering
\caption{Experiment2-LUBM with simple meta-queries (execution times in seconds)}
\label{ex2lubmMtable}
\begin{tabular}{|c|c|c|c|c|} 
\noalign{\hrule height 0.5pt}
                   & \textbf{mq1}   & \textbf{mq4}   & \textbf{mq5}   & \textbf{mq10}   \\ 
\noalign{\hrule height 0.5pt}
\multicolumn{5}{|c|}{\textbf{LUBM(1)}}  \\ 
\noalign{\hrule height 0.5pt}
\textbf{LogicBlox} & 93.95 & 92.21 & 92.26 & 93.09  \\ \noalign{\hrule height 0.5pt}
        \textbf{RDFox} & 2.38 & 2.41 & 2.37 & 2.59  \\ \noalign{\hrule height 0.5pt}
        \textbf{XSB} & 0.19 & 0 & 0.07 & 0.07  \\ \noalign{\hrule height 0.5pt}
        \textbf{NoHR} & 64.28 & 52.37 & 55.46 & 59.05  \\ \noalign{\hrule height 0.5pt}
        \textbf{Clingo} & \textbf{0.11} & \textbf{0.1} & \textbf{0.11} & \textbf{0.1}  \\ \noalign{\hrule height 0.5pt}
        \textbf{DLV2} & 0.12 & 0.12 & 0.12 & 0.11  \\ \noalign{\hrule height 0.5pt}
        \textbf{DLVHex} & 9.89 & 7.26 & 7.83 & 7.24  \\ \noalign{\hrule height 0.5pt}
        \textbf{HexLite} & OOT & 2.94 & OOT & 101.56  \\ \noalign{\hrule height 0.5pt}
        \textbf{Alpha} & 101.05 & 90.95 & 95.99 & 95.55  \\ \noalign{\hrule height 0.5pt}
\multicolumn{5}{|c|}{\textbf{LUBM(9)}}  \\ 
\noalign{\hrule height 0.5pt}
        \textbf{LogicBlox} & 1122.21 & 1156.2 & 1168.95 & 1140.84  \\ \noalign{\hrule height 0.5pt}
        \textbf{RDFox} & 49.54 & 49.6 & 49.54 & 49.51  \\ \noalign{\hrule height 0.5pt}
        \textbf{XSB} & 17.16 & 16.62 & 18.21 & 16.63  \\ \noalign{\hrule height 0.5pt}
        \textbf{NoHR} & OOT & OOT & OOT & OOT  \\ \noalign{\hrule height 0.5pt}
        \textbf{Clingo} & 1.33 & 1.32 & 1.28 & 1.32  \\ \noalign{\hrule height 0.5pt}
        \textbf{DLV2} & \textbf{1.03} & \textbf{0.97} & \textbf{1.01} & \textbf{0.97}  \\ \noalign{\hrule height 0.5pt}
        \textbf{DLVHex} & 539.97 & 385.85 & 410.25 & 387.26  \\ \noalign{\hrule height 0.5pt}
        \textbf{HexLite} & OOT & 635.33 & OOT & OOT  \\ \noalign{\hrule height 0.5pt}
        \textbf{Alpha} & 556.62 & 596.96 & 576.99 & 535.58  \\ \noalign{\hrule height 0.5pt}
\noalign{\hrule height 0.5pt}
\end{tabular}
\end{table}

\begin{table}
\centering
\caption{Experiment2-LUBM-ext with meta-queries (execution times in seconds)}
\label{ex2lubmMtable}
\begin{tabular}{|c|c|c|} 
\noalign{\hrule height 0.5pt}
                   & \textbf{sq1}   & \textbf{sq2}    \\ 
\noalign{\hrule height 0.5pt}
\multicolumn{3}{|c|}{\textbf{LUBM(1)-ext}}                \\ 
\noalign{\hrule height 0.5pt}
\textbf{LogicBlox} & 92.84 & 91.88  \\ \noalign{\hrule height 0.5pt}
        \textbf{RDFox} & 2.38 & 2.38  \\ \noalign{\hrule height 0.5pt}
        \textbf{XSB} & 0.07 & 0.07  \\ \noalign{\hrule height 0.5pt}
        \textbf{NoHR} & 57 & 56.4  \\ \noalign{\hrule height 0.5pt}
        \textbf{Clingo} & 0.11 & 0.11  \\ \noalign{\hrule height 0.5pt}
        \textbf{DLV2} & \textbf{0.1} & \textbf{0.11}  \\ \noalign{\hrule height 0.5pt}
        \textbf{DLVHex} & 7.78 & 7.79  \\ \noalign{\hrule height 0.5pt}
        \textbf{HexLite} & 243.85 & 50.93  \\ \noalign{\hrule height 0.5pt}
        \textbf{Alpha} & 92.91 & 94.21  \\ \noalign{\hrule height 0.5pt}
\multicolumn{3}{|c|}{\textbf{LUBM(9)-ext}}                \\ 
\noalign{\hrule height 0.5pt}
        \textbf{LogicBlox} & 1138.65 & 1074.2  \\ \noalign{\hrule height 0.5pt}
        \textbf{RDFox} & 49.41 & 49.56  \\ \noalign{\hrule height 0.5pt}
        \textbf{XSB} & 17.15 & 17.16  \\ \noalign{\hrule height 0.5pt}
        \textbf{NoHR} & OOT & OOT  \\ \noalign{\hrule height 0.5pt}
        \textbf{Clingo} & 1.33 & 1.33  \\ \noalign{\hrule height 0.5pt}
        \textbf{DLV2} & \textbf{0.97} & \textbf{1.01}  \\ \noalign{\hrule height 0.5pt}
        \textbf{DLVHex} & 700.97 & 706.53  \\ \noalign{\hrule height 0.5pt}
        \textbf{HexLite} & OOT & 1591.73  \\ \noalign{\hrule height 0.5pt}
        \textbf{Alpha} & 530.09 & 533.14  \\ \noalign{\hrule height 0.5pt}
\noalign{\hrule height 0.5pt}
\end{tabular}
\end{table}

\begin{table}[h!]
\centering
\caption{Experiment2-MODEUS with meta-queries (execution times in seconds)}
\label{ex2modustable}
\scalebox{1.0}{
\begin{tabular}{*{9}{|@{}c@{}}|}
\noalign{\hrule height 0.5pt}
\multicolumn{1}{|l|}{}                   & \multicolumn{1}{l|}{\textbf{mq0}} & \multicolumn{1}{l|}{\textbf{mq1}} & \multicolumn{1}{l|}{\textbf{mq2}} & \multicolumn{1}{l|}{\textbf{mq3}} & \multicolumn{1}{l|}{\textbf{mq4}} & \multicolumn{1}{l|}{\textbf{mq5}} & \multicolumn{1}{l|}{\textbf{mq6}} & \textbf{mq8} \\ \noalign{\hrule height 0.5pt}
\multicolumn{9}{|c|}{\textbf{MEF-00}}                                                                                 \\ \noalign{\hrule height 0.5pt}
\multicolumn{1}{|l|}{\textbf{LogicBlox}} & \multicolumn{1}{l|}{OOT}          & \multicolumn{1}{l|}{OOT}          & \multicolumn{1}{l|}{OOT}          & \multicolumn{1}{l|}{OOT}          & \multicolumn{1}{l|}{OOT}          & \multicolumn{1}{l|}{OOT}          & \multicolumn{1}{l|}{OOT}          & OOT          \\ \noalign{\hrule height 0.5pt}
\multicolumn{1}{|l|}{\textbf{RDFox}}     & \multicolumn{1}{l|}{1575.92}          & \multicolumn{1}{l|}{1561.46 }          & \multicolumn{1}{l|}{1578.32}          & \multicolumn{1}{l|}{1559.66}          & \multicolumn{1}{l|}{1665.68}          & \multicolumn{1}{l|}{1583.01}          & \multicolumn{1}{l|}{1579.26}          & 1565.73          \\ \noalign{\hrule height 0.5pt}
\multicolumn{1}{|l|}{\textbf{XSB}}       & 37.92 & 42.36 & 48.6 & 44.15 & 38.98 & 52.45 & 72.22 & 37.96       \\ \noalign{\hrule height 0.5pt}
\multicolumn{1}{|l|}{\textbf{NoHR}}      & 550.47 & 1040.42 & OOT & 595.65 & 224.36 & 1687.22 & 460.85 & 206.07       \\ \noalign{\hrule height 0.5pt}
\multicolumn{1}{|l|}{\textbf{Clingo}}    & \multicolumn{1}{l|}{OOT}          & \multicolumn{1}{l|}{OOT}          & \multicolumn{1}{l|}{OOT}          & \multicolumn{1}{l|}{OOT}          & \multicolumn{1}{l|}{OOT}          & \multicolumn{1}{l|}{OOT}          & \multicolumn{1}{l|}{OOT}          & OOT          \\ \noalign{\hrule height 0.5pt}
\multicolumn{1}{|l|}{\textbf{DLV2}}       & \multicolumn{1}{l|}{\textbf{5.02}}        & \multicolumn{1}{l|}{\textbf{5.21}}        & \multicolumn{1}{l|}{\textbf{7.13}}        & \multicolumn{1}{l|}{\textbf{5.07}}        & \multicolumn{1}{l|}{\textbf{4.29}}        & \multicolumn{1}{l|}{\textbf{4.29}}        & \multicolumn{1}{l|}{\textbf{4.33}}    &\multicolumn{1}{l|}{\textbf{ 4.2}}        \\ \noalign{\hrule height 0.5pt}
\multicolumn{1}{|l|}{\textbf{DLVHex}}    & \multicolumn{1}{l|}{OOT}          & \multicolumn{1}{l|}{OOT}          & \multicolumn{1}{l|}{OOT}          & \multicolumn{1}{l|}{OOT}          & \multicolumn{1}{l|}{OOT}          & \multicolumn{1}{l|}{OOT}          & \multicolumn{1}{l|}{OOT}          & OOT          \\ \noalign{\hrule height 0.5pt}
\multicolumn{1}{|l|}{\textbf{HexLite}}   & \multicolumn{1}{l|}{OOM}          & \multicolumn{1}{l|}{OOM}          & \multicolumn{1}{l|}{OOM}          & \multicolumn{1}{l|}{OOM}          & \multicolumn{1}{l|}{OOM}          & \multicolumn{1}{l|}{OOM}          & \multicolumn{1}{l|}{OOM}          & OOM          \\ \noalign{\hrule height 0.5pt}
\multicolumn{1}{|l|}{\textbf{Alpha}}     & \multicolumn{1}{l|}{OOT}          & \multicolumn{1}{l|}{OOT}          & \multicolumn{1}{l|}{OOT}          & \multicolumn{1}{l|}{OOT}          & \multicolumn{1}{l|}{OOT}          & \multicolumn{1}{l|}{OOT}          & \multicolumn{1}{l|}{OOT}          & OOT          \\ \noalign{\hrule height 0.5pt}
\multicolumn{9}{|c|}{\textbf{MEF-01}}                                                                 \\ \noalign{\hrule height 0.5pt}
\multicolumn{1}{|l|}{\textbf{LogicBlox}} & \multicolumn{1}{l|}{OOT}          & \multicolumn{1}{l|}{OOT}          & \multicolumn{1}{l|}{OOT}          & \multicolumn{1}{l|}{OOT}          & \multicolumn{1}{l|}{OOT}          & \multicolumn{1}{l|}{OOT}          & \multicolumn{1}{l|}{OOT}          & OOT          \\ \noalign{\hrule height 0.5pt}
\multicolumn{1}{|l|}{\textbf{RDFox}}     & \multicolumn{1}{l|}{OOT}          & \multicolumn{1}{l|}{OOT}          & \multicolumn{1}{l|}{OOT}          & \multicolumn{1}{l|}{OOT}          & \multicolumn{1}{l|}{OOT}          & \multicolumn{1}{l|}{OOT}          & \multicolumn{1}{l|}{OOT}          & OOT          \\ \noalign{\hrule height 0.5pt}
\multicolumn{1}{|l|}{\textbf{XSB}}       & 76.57 & 81.59 & 89.98 & 83.6 & 77.42 & 93.77 & 119.21 & 77.44        \\ \noalign{\hrule height 0.5pt}
\multicolumn{1}{|l|}{\textbf{NoHR}}      & 702.06 & 1697.3 & OOT & 1176.21 & 321.57 & OOT & 412.93 & 456.43      \\ \noalign{\hrule height 0.5pt}
\multicolumn{1}{|l|}{\textbf{Clingo}}    & \multicolumn{1}{l|}{OOT}          & \multicolumn{1}{l|}{OOT}          & \multicolumn{1}{l|}{OOT}          & \multicolumn{1}{l|}{OOT}          & \multicolumn{1}{l|}{OOT}          & \multicolumn{1}{l|}{OOT}          & \multicolumn{1}{l|}{OOT}          & OOT          \\ \noalign{\hrule height 0.5pt}
\multicolumn{1}{|l|}{\textbf{DLV2}}       & \multicolumn{1}{l|}{\textbf{6.19}}        & \multicolumn{1}{l|}{\textbf{6.23}}        & \multicolumn{1}{l|}{\textbf{8.27}}        & \multicolumn{1}{l|}{\textbf{6.35}}        & \multicolumn{1}{l|}{\textbf{5.46}}        & \multicolumn{1}{l|}{\textbf{6.8}}        & \multicolumn{1}{l|}{\textbf{5.38}}        & \textbf{5.3}        \\ \noalign{\hrule height 0.5pt}
\multicolumn{1}{|l|}{\textbf{DLVHex}}    & \multicolumn{1}{l|}{OOT}          & \multicolumn{1}{l|}{OOT}          & \multicolumn{1}{l|}{OOT}          & \multicolumn{1}{l|}{OOT}          & \multicolumn{1}{l|}{OOT}          & \multicolumn{1}{l|}{OOT}          & \multicolumn{1}{l|}{OOT}          & OOT          \\ \noalign{\hrule height 0.5pt}
\multicolumn{1}{|l|}{\textbf{HexLite}}   & \multicolumn{1}{l|}{OOM}          & \multicolumn{1}{l|}{OOM}          & \multicolumn{1}{l|}{OOM}          & \multicolumn{1}{l|}{OOM}          & \multicolumn{1}{l|}{OOM}          & \multicolumn{1}{l|}{OOM}          & \multicolumn{1}{l|}{OOM}          & OOM          \\ \noalign{\hrule height 0.5pt}
\multicolumn{1}{|l|}{\textbf{Alpha}}     & \multicolumn{1}{l|}{OOT}          & \multicolumn{1}{l|}{OOT}          & \multicolumn{1}{l|}{OOT}          & \multicolumn{1}{l|}{OOT}          & \multicolumn{1}{l|}{OOT}          & \multicolumn{1}{l|}{OOT}          & \multicolumn{1}{l|}{OOT}          & OOT          \\ \noalign{\hrule height 0.5pt}
\multicolumn{9}{|c|}{\textbf{MEF-02}}                                                                                                    \\ \noalign{\hrule height 0.5pt}
\multicolumn{1}{|l|}{\textbf{LogicBlox}} & \multicolumn{1}{l|}{OOT}          & \multicolumn{1}{l|}{OOT}          & \multicolumn{1}{l|}{OOT}          & \multicolumn{1}{l|}{OOT}          & \multicolumn{1}{l|}{OOT}          & \multicolumn{1}{l|}{OOT}          & \multicolumn{1}{l|}{OOT}          & OOT          \\ \noalign{\hrule height 0.5pt}
\multicolumn{1}{|l|}{\textbf{RDFox}}     & \multicolumn{1}{l|}{1570.9}          & \multicolumn{1}{l|}{1578.77}          & \multicolumn{1}{l|}{1607.26}          & \multicolumn{1}{l|}{1565.8}          & \multicolumn{1}{l|}{1570.86}          & \multicolumn{1}{l|}{1626.46}          & \multicolumn{1}{l|}{1622.68}          & 1574.04          \\ \noalign{\hrule height 0.5pt}
\multicolumn{1}{|l|}{\textbf{XSB}}       & 38.83 & 41.04 & 48.25 & 44.18 & 39 & 50.39 & 74.31 & 38.04       \\ \noalign{\hrule height 0.5pt}
\multicolumn{1}{|l|}{\textbf{NoHR}}      & 561.18 & 868.59 & OOT & 966.51 & 219.15 & 1057.22 & 311.15 & 285.47      \\ \noalign{\hrule height 0.5pt}
\multicolumn{1}{|l|}{\textbf{Clingo}}    & \multicolumn{1}{l|}{OOT}          & \multicolumn{1}{l|}{OOT}          & \multicolumn{1}{l|}{OOT}          & \multicolumn{1}{l|}{OOT}          & \multicolumn{1}{l|}{OOT}          & \multicolumn{1}{l|}{OOT}          & \multicolumn{1}{l|}{OOT}          & OOT          \\ \noalign{\hrule height 0.5pt}
\multicolumn{1}{|l|}{\textbf{DLV2}}       & \multicolumn{1}{l|}{\textbf{5.01}}        & \multicolumn{1}{l|}{\textbf{5.49}}        & \multicolumn{1}{l|}{\textbf{7.16}}        & \multicolumn{1}{l|}{\textbf{5.03}}        & \multicolumn{1}{l|}{\textbf{4.26}}        & \multicolumn{1}{l|}{\textbf{4.24}}        & \multicolumn{1}{l|}{\textbf{4.640}}        & \textbf{4.590}        \\ \noalign{\hrule height 0.5pt}
\multicolumn{1}{|l|}{\textbf{DLVHex}}    & \multicolumn{1}{l|}{OOT}          & \multicolumn{1}{l|}{OOT}          & \multicolumn{1}{l|}{OOT}          & \multicolumn{1}{l|}{OOT}          & \multicolumn{1}{l|}{OOT}          & \multicolumn{1}{l|}{OOT}          & \multicolumn{1}{l|}{OOT}          & OOT          \\ \noalign{\hrule height 0.5pt}
\multicolumn{1}{|l|}{\textbf{HexLite}}   & \multicolumn{1}{l|}{OOM}          & \multicolumn{1}{l|}{OOM}          & \multicolumn{1}{l|}{OOM}          & \multicolumn{1}{l|}{OOM}          & \multicolumn{1}{l|}{OOM}          & \multicolumn{1}{l|}{OOM}          & \multicolumn{1}{l|}{OOM}          & OOM          \\ \noalign{\hrule height 0.5pt}
\multicolumn{1}{|l|}{\textbf{Alpha}}     & \multicolumn{1}{l|}{OOT}          & \multicolumn{1}{l|}{OOT}          & \multicolumn{1}{l|}{OOT}          & \multicolumn{1}{l|}{OOT}          & \multicolumn{1}{l|}{OOT}          & \multicolumn{1}{l|}{OOT}          & \multicolumn{1}{l|}{OOT}          & OOT          \\ \noalign{\hrule height 0.5pt}
\multicolumn{9}{|c|}{\textbf{MEF-03}}                                                                                           \\ \noalign{\hrule height 0.5pt}
\multicolumn{1}{|l|}{\textbf{LogicBlox}} & \multicolumn{1}{l|}{OOT}          & \multicolumn{1}{l|}{OOT}          & \multicolumn{1}{l|}{OOT}          & \multicolumn{1}{l|}{OOT}          & \multicolumn{1}{l|}{OOT}          & \multicolumn{1}{l|}{OOT}          & \multicolumn{1}{l|}{OOT}          & OOT          \\ \noalign{\hrule height 0.5pt}
\multicolumn{1}{|l|}{\textbf{RDFox}}     & \multicolumn{1}{l|}{OOT}          & \multicolumn{1}{l|}{OOT}          & \multicolumn{1}{l|}{OOT}          & \multicolumn{1}{l|}{OOT}          & \multicolumn{1}{l|}{OOT}          & \multicolumn{1}{l|}{OOT}          & \multicolumn{1}{l|}{OOT}          & OOT          \\ \noalign{\hrule height 0.5pt}
\multicolumn{1}{|l|}{\textbf{XSB}}       & 76.43 & 80.58 & 88.88 & 87.43 & 77.46 & 91.01 & 121.58 & 76.29       \\ \noalign{\hrule height 0.5pt}
\multicolumn{1}{|l|}{\textbf{NoHR}}      & 671.79 & 1029.06 & OOT & 689.84 & 431.93 & 1614.51 & 389.34 & 428.02      \\ \noalign{\hrule height 0.5pt}
\multicolumn{1}{|l|}{\textbf{Clingo}}    & \multicolumn{1}{l|}{OOT}          & \multicolumn{1}{l|}{OOT}          & \multicolumn{1}{l|}{OOT}          & \multicolumn{1}{l|}{OOT}          & \multicolumn{1}{l|}{OOT}          & \multicolumn{1}{l|}{OOT}          & \multicolumn{1}{l|}{OOT}          & OOT          \\ \noalign{\hrule height 0.5pt}
\multicolumn{1}{|l|}{\textbf{DLV2}}       & \multicolumn{1}{l|}{\textbf{5.74}}        & \multicolumn{1}{l|}{\textbf{6.71}}        & \multicolumn{1}{l|}{\textbf{8.79}}       & \multicolumn{1}{l|}{\textbf{5.84}}        & \multicolumn{1}{l|}{\textbf{5.23}}        & \multicolumn{1}{l|}{\textbf{6.62}}        & \multicolumn{1}{l|}{\textbf{5.18}}        & \textbf{5.06}        \\ \noalign{\hrule height 0.5pt}
\multicolumn{1}{|l|}{\textbf{DLVHex}}    & \multicolumn{1}{l|}{OOT}          & \multicolumn{1}{l|}{OOT}          & \multicolumn{1}{l|}{OOT}          & \multicolumn{1}{l|}{OOT}          & \multicolumn{1}{l|}{OOT}          & \multicolumn{1}{l|}{OOT}          & \multicolumn{1}{l|}{OOT}          &     OOT      \\ \noalign{\hrule height 0.5pt}
\multicolumn{1}{|l|}{\textbf{HexLite}}   & \multicolumn{1}{l|}{OOM}          & \multicolumn{1}{l|}{OOM}          & \multicolumn{1}{l|}{OOM}          & \multicolumn{1}{l|}{OOM}          & \multicolumn{1}{l|}{OOM}          & \multicolumn{1}{l|}{OOM}          & \multicolumn{1}{l|}{OOM}          & OOM          \\ \noalign{\hrule height 0.5pt}
\multicolumn{1}{|l|}{\textbf{Alpha}}     & \multicolumn{1}{l|}{OOT}          & \multicolumn{1}{l|}{OOT}          & \multicolumn{1}{l|}{OOT}          & \multicolumn{1}{l|}{OOT}          & \multicolumn{1}{l|}{OOT}          & \multicolumn{1}{l|}{OOT}          & \multicolumn{1}{l|}{OOT}          &      OOT     \\ \noalign{\hrule height 0.5pt}
\end{tabular}}
\end{table}

\end{document}